\documentclass[10pt]{article} %
\usepackage[preprint]{tmlr}

\usepackage{amsmath,amsfonts,bm}

\def\eqref#1{equation~\ref{#1}}

\def\1{\bm{1}}

\DeclareMathAlphabet{\mathsfit}{\encodingdefault}{\sfdefault}{m}{sl}
\SetMathAlphabet{\mathsfit}{bold}{\encodingdefault}{\sfdefault}{bx}{n}

\usepackage{enumitem}
\usepackage{hyperref}
\usepackage{url}
\usepackage{booktabs}
\usepackage{makecell}
\usepackage{graphicx}
\usepackage{xcolor}
\usepackage{listings}
\usepackage{mathrsfs}
\usepackage{amssymb}
\usepackage[most]{tcolorbox}
\tcbuselibrary{skins, breakable}
\colorlet{lightblue}{blue!10!white}

\definecolor{contriborange}{HTML}{ee9c39}
\definecolor{contribblue}{HTML}{005CA0}      %

\newtcolorbox{contribbox}[1][]{
  enhanced,
  title=#1,
  colback=contriborange!10,
  colbacktitle=contriborange!10,
  coltitle=black,
  left=0pt, right=8pt, top=10pt, bottom=2pt,
  attach boxed title to top left={xshift=15pt,yshift=-10pt},
  boxed title style={frame hidden,colback=contriborange!45},
  sharp corners, rounded corners, arc=3pt
}

\definecolor{codebg}{RGB}{247,247,250}
\definecolor{codekw}{RGB}{0,92,160}      %
\definecolor{codecomment}{RGB}{120,120,120}
\definecolor{codestring}{RGB}{140,90,40}
\definecolor{hypaccent}{RGB}{130,40,150} %

\lstdefinestyle{pystyle}{
    language=Python,
    backgroundcolor=\color{codebg},
    basicstyle=\ttfamily\scriptsize,
    keywordstyle=\color{codekw}\bfseries,
    commentstyle=\color{codecomment}\itshape,
    stringstyle=\color{codestring},
    showstringspaces=false,
    breaklines=true,
    frame=single,
    rulecolor=\color{black!30},
    framesep=2pt,
    xleftmargin=0pt,
    tabsize=4,
    emph={Hyperboloid,HTCLinear,HypRegressionHyperboloid,hrc_relu,expmap_0},
    emphstyle=\color{hypaccent},
}

\def\month{MM}
\def\year{2026}

\title{hyperbolix: Hyperbolic Deep Learning in JAX}

\author{\name Timo Klein \email timo.klein@univie.ac.at \\
      \addr Faculty of Computer Science, University of Vienna, Vienna, Austria \\
      UniVie Doctoral School Computer Science, University of Vienna, Vienna, Austria
      \AND
      \name Thomas Lang \email thomas.lang@univie.ac.at \\
      \addr Faculty of Computer Science, University of Vienna, Vienna, Austria\\
      UniVie Doctoral School Computer Science, University of Vienna, Vienna, Austria
      \AND
      \name Yllka Velaj \email yllka.velaj@univie.ac.at \\
      \addr Faculty of Computer Science, University of Vienna, Vienna, Austria\\
      ds:UniVie, University of Vienna, Vienna, Austria
      \AND
      \name Sebastian Tschiatschek \email sebastian.tschiatschek@univie.ac.at \\
      \addr Faculty of Computer Science, University of Vienna, Vienna, Austria\\
      ds:UniVie, University of Vienna, Vienna, Austria
    }

\begin{document}

\maketitle

\begin{abstract}
We present \textbf{hyperbolix}, an open-source library for hyperbolic deep learning in JAX, built on Flax NNX. To our knowledge, it is the first comprehensive, general-purpose hyperbolic deep learning library in JAX. It includes six manifolds with a common interface: Euclidean space, the Poincaré ball, the hyperboloid, the $\kappa$-stereographic model, mixed-curvature product spaces, and the proper velocity space. We implement layer families that cover linear layers, convolutions, attention, normalization, positional encoding, regression, and vector quantization. These building blocks span methods ranging from Ganea's original hyperbolic neural networks to recent fully hyperbolic architectures such as Hypformer and Lorentzian ResNet. Additionally, hyperbolix contains Riemannian optimizers implemented as optax transformations, wrapped distributions, and hyperbolic dimensionality-reduction techniques. Its API uses idiomatic JAX: Manifolds are stateless, with curvature being passed at call time, while manifold operations act on single points, with \texttt{jax.vmap} enabling batch operations. The precision of every checked operation is tested against a closed-form NumPy/SciPy transcription from the source paper or a finite difference, for both float32 and float64. On the hyperboloid, standard formulas for two-point operations, such as the distance, lose precision far from the origin, because they subtract two large, nearly equal terms. hyperbolix replaces these subtractions with cancellation-free formulas that stay accurate in float32 at distances where prior implementations return NaN. hyperbolix is available under the MIT license at \url{https://github.com/timoklein/hyperbolix}.
\end{abstract}

\section{Introduction}
\label{sec:intro}

Deep learning reaches practitioners through frameworks such as PyTorch~\citep{paszke2019pytorch} or the JAX ecosystem~\citep{bradbury2018jax,heek2020flax}. Operations readily available in these frameworks, in turn, shape the methods researchers develop. Both PyTorch and JAX/Flax are Euclidean by construction: tensors live in $\mathbb{R}^n$, and common deep learning components such as linear layers, residual connections, and optimizer steps rely on Euclidean operations like affine transformations and vector addition. The assumption that data lives in a vector space is rarely made explicit because it rarely constrains what can be modeled. Hierarchical data violates that assumption. Examples include natural language, where words fall into semantic categories, and taxonomic classification, where each class is embedded within a broader superclass. Both impose a tree structure that a flat vector space cannot represent efficiently. The number of nodes in a hierarchy grows exponentially with depth. Hyperbolic space, a Riemannian manifold of constant negative curvature, matches this growth rate: its volume expands exponentially with radius, making it the natural geometry for tree-like data \citep{ungar2009gyrovector, Sarkar2011LowDistortion}. But hyperbolic space is not a vector space, so operations like vector addition do not transfer directly. Hyperbolic deep learning has had to rebuild them: \citet{ganea2018hnn} introduced the first hyperbolic linear layer and softmax classifier, and the family has grown to include reformulated layers~\citep{shimizu2021hnnpp, shi2026ilnn, vanDerKlis2026fgglnn}, convolutional backbones~\citep{bdeir2024fhcnn,vanspengler2023resnet, he2025lresnet}, transformers and self-attention~\citep{chen2022fhnn, yang2025hypformer, he2025hope_helm}, vision\text{--}language models~\citep{pal2025hycocplip, yoshikawa2026phyclip}, as well as mixed-curvature product spaces~\citep{gu2019mixed, skopek2020mixedcurvaturevae}. A more thorough overview of current architectures and trends can be found in the numerous surveys discussing the literature, e.g., \citet{mettes2024hypcvsurvey} or \citet{he2025hypfoundationsurvey}.

Which of these methods are readily available to practitioners depends on the framework they use. In PyTorch, geoopt~\citep{kochurov2020geoopt} is the most comprehensive library, containing manifolds, neural network layers, and Riemannian optimizers. HypLL~\citep{vanspengler2023hypll} focuses on the Poincar\'e ball, providing hyperbolic layers. Geomstats~\citep{miolane2020geomstats} covers general Riemannian geometry across backends but is not a deep learning library. JAX has Riemannian optimizers too: Rieoptax \citep{utpala2022rieoptax}, RiemannAX~\citep{mary2024riemannax}, and GeoJAX \citep{you2026geojax} all provide them, but none offer a general-purpose hyperbolic layer library. Rieoptax contains a handful of undocumented Poincar\'e and Lorentz layers but has been inactive since 2023, while GeoJAX and RiemannAX ship no hyperbolic layers at all. Thus, the only option for JAX users in need of hyperbolic linear layers, attention blocks, or normalization layers is to implement them from scratch or to hand-port an existing implementation. However, such a port is a tricky endeavor due to common numerical stability issues. 

To close this gap in the JAX ecosystem, we present hyperbolix: a hyperbolic deep learning library for JAX~\citep{bradbury2018jax} and Flax NNX~\citep{heek2020flax}. It implements six manifolds with a unified interface: three common models of hyperbolic geometry, the Poincar\'e ball, the hyperboloid, and the proper velocity model, alongside a signed-curvature $\kappa$-stereographic model \citep{bachmann2020kappa_stereographic}, Euclidean space, and mixed-curvature manifolds~\citep{gu2019mixed}. Exact isometries connect the three hyperbolic models, so a representation learned in one transfers to another without retraining. Our \texttt{nn\_layers} package provides 76 implementations that cover the linear, convolution, attention, normalization, positional encoding, regression, and vector quantization layers of standard architectures. Around the layers, we also implement Riemannian Adam and SGD~\citep{bonnabel2013rsgd,becigneul2018riemannian} written as optax transformations~\citep{deepmind2020optax}, wrapped normal distributions~\citep{Nagano2019wrapped_normal}, as well as HoroPCA \citep{chami2021horopca} and CO-SNE \citep{guo2022cosne} as dimensionality reduction techniques. hyperbolix is available under the MIT license at \url{https://github.com/timoklein/hyperbolix}.

\begin{contribbox}[\textbf{Key Contributions}]
\begin{enumerate}[leftmargin=0.7cm]
\item \textbf{\textcolor{contribblue}{The library.}} To our knowledge, hyperbolix is the first comprehensive, general-purpose hyperbolic deep learning library for JAX (Section~\ref{sec:contents}, Table~\ref{tab:comparison}). Its API adapts manifold operations to JAX's functional paradigm, illustrated by a worked example that converts a Euclidean model to its hyperbolic counterpart (Section~\ref{sec:design}, Figure~\ref{fig:code}).
\item \textbf{\textcolor{contribblue}{Systematic correctness verification.}} Every checked operation is compared against an independently derived reference rather than another library call (Section~\ref{sec:correctness}).
\item \textbf{\textcolor{contribblue}{Higher numerical accuracy than existing libraries.}} On the Poincar\'e ball, hyperbolix matches or exceeds the round-trip precision of geoopt and HypLL. On the hyperboloid, we replace every Minkowski inner-product subtraction with a cancellation-free formula~\citep{Blasius2016EfficientEmbeddings, Celinska2024NumericalAspectsHyp}, staying at float32 rounding level up to geodesic radius 10, where geoopt already returns NaN (Section~\ref{sec:stability}, Figures~\ref{fig:cancellation_accuracy} and~\ref{fig:crosslib_stability}).
\end{enumerate}
\end{contribbox}

\section{Design and API}
\label{sec:design}

This section presents the API of hyperbolix, starting with the design principles that make it compatible with JAX transformations. Next, we introduce learnable curvature, Riemannian optimization, and initialization.

\subsection{Core design principles}

Every manifold operation in hyperbolix is a pure function, meaning its output depends only on its inputs, and it modifies no external state. This makes hyperbolix composable with \texttt{jax.jit}, \texttt{jax.vmap}, and \texttt{jax.grad}~\citep{bradbury2018jax}, implemented through the following three design decisions.

\emph{Manifold operations take unbatched inputs.} A distance takes two arrays of shape \texttt{(dim,)} and returns a scalar; $\exp_0$ takes a tangent vector and returns a manifold point, again with no batch axis. Batching is an explicit \texttt{jax.vmap} written by the caller, who has to state mapped axes. A batched implementation would fix the layout of an operation, requiring callers to reshape inputs for pairwise, broadcast, and per-example uses of the same distance. The unbatched design leaves the choice of layout to the caller.

\emph{Manifolds are plain Python classes, not \texttt{nnx.Module}s.} Because curvature is passed into every call, a manifold instance holds no NNX state, and Flax NNX does not track it as part of the module graph. One instance can therefore be shared across all layers, including those within \texttt{nnx.scan} or \texttt{nnx.fori\_loop}, which reject objects whose state is referenced from multiple places.

\emph{Curvature is passed at call time.} We write $c$ for the curvature parameter of a hyperbolic manifold, which has constant sectional curvature $-c$ for $c > 0$. The $\kappa$-stereographic manifold takes a signed $\kappa$ whose sign selects hyperbolic, Euclidean, or spherical geometry~\citep{bachmann2020kappa_stereographic}. Because $c$ is an argument rather than a construction-time state, it can be a Python float (a compile-time constant under \texttt{jit}) or a traced \texttt{jax.Array}, and the same manifold instance serves both fixed and learned curvature.

Figure~\ref{fig:code} shows a Euclidean Flax NNX model beside its hyperboloid counterpart using hyperbolix. Three key differences are visible. First, the Euclidean features are lifted onto the manifold in \texttt{\_\_call\_\_}. Prepending a zero time coordinate maps them to the tangent space at the origin, and $\exp_0$ sends the result to the hyperboloid under an explicit \texttt{jax.vmap}. Second, each \texttt{HTCLinear} layer~\citep{yang2025hypformer} applies a Euclidean linear transformation to the input point and then recomputes the time coordinate, ensuring that the output lies on the hyperboloid. The layer optionally accepts separate input and output curvatures, enabling a model to change curvature between layers. Third, the classifier is a hyperbolic multinomial logistic regression head that uses points on the hyperboloid as inputs. The model class, parameter handling, and training loop are unchanged from the Euclidean version, so these three changes are the full cost of the conversion.

\subsection{Learnable curvature}

Curvature is often better treated as a parameter than a hyperparameter. \texttt{LearnableCurvature} is a standalone \texttt{nnx.Module} held by the model that yields $c$ on call. The manifold itself stays stateless. \texttt{LearnableCurvature} stores a single real-valued scalar corresponding to the curvature through one of three parameterizations. The default, \texttt{log}, sets $c = \exp(r)$ for an unconstrained parameter $r$. Each step in $r$ then changes $c$ by a constant factor rather than by a constant amount, which suits curvatures spanning orders of magnitude~\citep{yoshikawa2026phyclip}. \texttt{softplus} sets $c = \mathrm{softplus}(r)$, guaranteeing $c > 0$ with derivative $\partial c/ \partial r = \sigma(r) \in (0,1)$~\citep{kochurov2020geoopt}. \texttt{identity} is the signed parameterization required for reaching the Euclidean and spherical values of the $\kappa$-stereographic manifold. The curvature is additionally optionally clamped to a fixed interval. Because $r$ is an unconstrained real number, a plain \texttt{optax.adam}~\citep{deepmind2020optax} can update it.

\subsection{Optimization}
\label{subsec:optim}

hyperbolix ships \texttt{riemannian\_adam} and \texttt{riemannian\_sgd} as \texttt{optax.GradientTransformation}s, so they compose with Optax and can plug into \texttt{nnx.Optimizer}. Manifold-valued parameters are detected automatically: \texttt{ManifoldParam}, an \texttt{nnx.Param} subclass, carries the manifold and curvature, and \texttt{mark\_manifold\_param} wraps an existing parameter. Every unmarked parameter receives the standard Euclidean update in the same transformation.

Most models avoid Riemannian SGD because modern layers parametrize manifold parameters as Euclidean parameters mapped onto the manifold~\citep{shimizu2021hnnpp,shi2026ilnn}. Their gradients are therefore Euclidean, making \texttt{optax.adam} the correct optimizer. Riemannian optimization is for parameters that genuinely live on the manifold: embedding tables optimized directly on the ball~\citep{nickel2017poincareembeddings, chen2025hvqvae} and legacy Ganea-style Poincar\'e layers whose bias is parameterized on the ball while the weight matrix stays Euclidean~\citep{ganea2018hnn}.

\subsection{Initialization}

Generic He or Xavier initialization fails in hyperbolic layers in two ways. If the initialization weights are too large, the layer's outputs land near the boundary of the Poincar\'e ball or far up the hyperboloid, where distances and their gradients become \texttt{NaN} within a few steps. If the weights are too small, the per-layer gain below one compounds as $\mathcal O \big(\text{gain}^{\text{depth}} \big)$ until the outputs of distinct inputs become numerically indistinguishable in the working dtype, collapsing the network to a constant function. Appendix~\ref{app:init} lists the default initialization of each layer family and its rationale, and Section~\ref{sec:stability} covers the underlying precision limits.

\begin{figure}[t]
  \centering
  \begin{minipage}[t]{0.48\textwidth}
    \begin{lstlisting}[style=pystyle]
import jax
from flax import nnx


class MLP(nnx.Module):
    def __init__(self, rngs: nnx.Rngs):
        self.embed = nnx.Linear(784, 32, rngs=rngs)
        self.fc1 = nnx.Linear(32, 64, rngs=rngs)
        self.fc2 = nnx.Linear(64, 64, rngs=rngs)
        self.head = nnx.Linear(64, 10, rngs=rngs)

    def __call__(self, x):
        x = jax.nn.relu(self.embed(x))
        x = jax.nn.relu(self.fc1(x))
        x = jax.nn.relu(self.fc2(x))
        return self.head(x)
    \end{lstlisting}
    \centering\small (a) Flax NNX (Euclidean)
  \end{minipage}
  \hfill
  \begin{minipage}[t]{0.48\textwidth}
    \begin{lstlisting}[style=pystyle]
import jax, jax.numpy as jnp
from flax import nnx
from hyperbolix.manifolds import Hyperboloid
from hyperbolix.nn_layers import (HTCLinear,
    HypRegressionHyperboloid, hrc_relu)

H = Hyperboloid()  # plain class, not a Module

class HypMLP(nnx.Module):
    def __init__(self, rngs: nnx.Rngs):
        self.embed = nnx.Linear(784, 32, rngs=rngs)
        self.fc1 = HTCLinear(33, 64, rngs=rngs)
        self.fc2 = HTCLinear(65, 64, rngs=rngs)
        self.head = HypRegressionHyperboloid(
            H, 65, 10, rngs=rngs, input_space="manifold")

    def __call__(self, x, c=1.0):
        x = jax.nn.relu(self.embed(x))
        x = jax.vmap(lambda v: H.expmap_0(
            jnp.concatenate([jnp.zeros(1), v]), c))(x)
        x = hrc_relu(self.fc1(x, c_in=c, c_out=c), c, c)
        x = hrc_relu(self.fc2(x, c_in=c, c_out=c), c, c)
        return self.head(x, c)
    \end{lstlisting}
    \centering\small (b) hyperbolix (hyperboloid)
  \end{minipage}
  \caption{\textbf{Converting a Euclidean MNIST MLP to a hybrid hyperbolic network.} Both models train with the same loop and a plain \texttt{optax.adam}, because both parameterize their weights in Euclidean space (Section~\ref{subsec:optim}). Three differences are visible in (b): the Euclidean features are projected onto the manifold via $\exp_0$ under an explicit \texttt{jax.vmap}, the curvature $c$ is passed as a call-time argument to every hyperbolic layer, and a hyperbolic regression head replaces the linear output layer. The dimensions 33 and 65 are ambient hyperboloid dimensions: 32 and 64 spatial coordinates, respectively, plus one time coordinate, which is fully determined by the space component.}
  \label{fig:code}
\end{figure}

\section{The hyperbolix Library}\label{sec:contents}

hyperbolix is built around six manifolds that share a common interface. The neural network layers, Riemannian optimizers, distributions, and dimensionality-reduction methods build on them. This section describes the manifolds first, then the layers, and the remaining packages. Finally, we compare hyperbolix to existing libraries.

\subsection{Manifolds}
 
All manifolds implement a common interface of 19 operations (Appendix~\ref{app:protocol}), so a user-defined manifold with the same methods works with the Riemannian optimizers and other manifold-generic components without subclassing any library class. The three hyperbolic models, the Poincar\'e ball, the hyperboloid, and the proper velocity space, pose different trade-offs. The hyperboloid yields larger gradient magnitudes than the Poincar\'e ball, aiding optimization~\citep{mishne2023numericalstability}, but its standard two-point formulas suffer catastrophic cancellation in the Minkowski inner product (Section~\ref{sec:stability}). The Poincar\'e ball is conformal to Euclidean space and supports 2D visualization, but its conformal factor diverges at the boundary resulting in unavoidable numerical issues. The proper velocity (PV) model represents points as unconstrained vectors in $\mathbb{R}^n$, avoiding $\exp_0$ and $\log_0$ projection steps~\citep{chen2026pvnn}. In hyperbolix, two-point operations (distance, parallel transport, gyrovector addition) leverage the fact that the PV $\leftrightarrow$ hyperboloid isometry is exact, allowing us to use the hyperboloid's cancellation-free arithmetic for PV operations (Section~\ref{sec:stability}). Exact isometries connect all three pairs of hyperbolic models (Poincar\'e $\leftrightarrow$ hyperboloid, proper velocity $\leftrightarrow$ Poincar\'e, proper velocity $\leftrightarrow$ hyperboloid), so a representation learned in one model transfers to another without retraining. The stereographic manifold~\citep{bachmann2020kappa_stereographic} unifies hyperbolic, Euclidean, and spherical geometry under one signed curvature. \citet{bachmann2020kappa_stereographic} parameterize this curvature as $\kappa$, with $\kappa<0$ hyperbolic, $\kappa=0$ Euclidean, and $\kappa>0$ spherical. We instead expose $c \doteq -\kappa$, so $c>0$ is hyperbolic, matching our Poincar\'e ball convention. Our manifold operations for the $\kappa$-stereographic model are analytically smooth at $c=0$ because the implementation evaluates their limits at $ c=0$ via Taylor expansions. Therefore, the gradient with respect to $c$ remains finite at the Euclidean point. With the \texttt{identity} parameterization of Section~\ref{sec:design}, learnable curvature can cross zero during training rather than being pinned to a fixed sign. \texttt{ProductManifold}~\citep{gu2019mixed} builds mixed-curvature product spaces from any of the above, takes curvature information as a per-factor tuple, and implements the same generic interface.

\begin{table}[t]
\centering
\caption{Contents of the \texttt{nn\_layers} package. A \checkmark{} marks manifold coverage. The full inventory with source attributions is in Appendix~\ref{app:layers}.}
\label{tab:layers}
\begin{tabular}{@{} l r ccc @{}}
\toprule
Slot & Variants & \textsc{Hyperboloid} & \textsc{Poincar\'e} & \textsc{Proper Velocity} \\
\midrule
Linear                      & 11 & \checkmark & \checkmark & \checkmark \\
Convolution                 &  7 & \checkmark & \checkmark & \checkmark \\
Attention                   &  4 & \checkmark &            &            \\
Normalization \& dropout    & 12 & \checkmark & \checkmark & \checkmark \\
Residual, pooling, primitives & 15 & \checkmark & \checkmark &          \\
Positional encoding         &  3 & \checkmark &            &            \\
Regression / MLR            &  7 & \checkmark & \checkmark & \checkmark \\
Vector quantization         &  3 &            & \checkmark &            \\
Activations                 & 13 & \checkmark & \checkmark &            \\
Interface\textsuperscript{\dag} &  1 & \checkmark & \checkmark &        \\
\midrule
Total                       & 76 \\
\bottomrule
\end{tabular}
 
\smallskip
{\footnotesize \textsuperscript{\dag} Hybrid regularizer (\texttt{HyperPPFeatureScaling}~\citep{klein2026hyperpp}; not itself on-manifold).}
\end{table}

\subsection{Neural network layers}
 
The \texttt{nn\_layers} package provides 44 layer classes, 31 functions, and one data type. Appendix~\ref{app:layers} lists the paper behind each. It contains linear, convolution, attention, normalization, positional-encoding, regression, activation, and vector quantization, comprising standard operations in deep learning. For a particular manifold and network layer, there can be multiple implementations with differing properties. This is for instance the case for the Poincaré linear layer: \citet{ganea2018hnn} parameterize it with a Euclidean weight matrix and a manifold bias, which requires Riemannian optimization. The HNN++ reformulation~\citep{shimizu2021hnnpp} moves all parameters off the manifold, so a standard \texttt{optax.adam} suffices. Similar parameterization trade-offs recur throughout the library: the convolution, attention, and normalization families each offer variants that target different manifolds and pose distinct trade-offs among fidelity, numerical stability, and speed. In appendices~\ref{app:layers} and~\ref{app:methods} we present the full list of layers and methods and where they were initially introduced.

\subsection{Beyond layers}
 
The \texttt{optim} package provides \texttt{riemannian\_adam} and \texttt{riemannian\_sgd}
as ordinary optax~\citep{deepmind2020optax} \texttt{GradientTransformation}s that detect manifold
parameters automatically. Section~\ref{subsec:optim} covers the implementation. 

We further implement three
distributions to support generative and probabilistic models: wrapped normals~\citep{Nagano2019wrapped_normal}
on the Poincar\'e ball and the hyperboloid, and a uniform distribution on the ball. 

For dimensionality reduction and visualization, \texttt{decomposition} implements HoroPCA~\citep{chami2021horopca} and CO-SNE~\citep{guo2022cosne}. 

\texttt{utils} exposes Gromov $\delta$-hyperbolicity estimation~\citep{khrulkov2020gromov_hyperbolicity}, which measures how tree-like a dataset is, together with the domain-clamped math primitives that Section~\ref{sec:stability} examines.

The documentation helps users choose among these implementations: for each layer family, such as linear or convolutional layers, a table states when to pick each layer. A separate numerical stability guide explains the failure modes of Section~\ref{sec:stability} and how to avoid them, for example, when to switch to float64.

\subsection{Comparison to existing libraries}
 
To our knowledge, hyperbolix is the first comprehensive, general-purpose \emph{hyperbolic deep
learning} library in JAX. While JAX already has Riemannian optimization
(Rieoptax~\citep{utpala2022rieoptax}, RiemannAX~\citep{mary2024riemannax},
GeoJAX~\citep{you2026geojax}), none of these libraries offers a general-purpose hyperbolic layer library. In PyTorch, geoopt~\citep{kochurov2020geoopt} and HypLL~\citep{vanspengler2023hypll} are
the closest counterparts; Geomstats~\citep{miolane2020geomstats} covers general Riemannian
geometry, but does not focus on deep learning. Table~\ref{tab:comparison} compares hyperbolix
against these libraries feature by feature. Appendix~\ref{app:jax} surveys the JAX libraries in detail.

   \begin{table}[t]
    \centering
    \small
    \setlength{\tabcolsep}{3pt}
    \caption{Feature comparison of hyperbolix and other existing manifold and hyperbolic deep learning
    libraries. P: Poincar\'e ball, H: hyperboloid, PV: proper velocity.
    Layer counts are the classes exported from each library's public neural network namespace.
    geoopt's released version exposes none. Geomstats is included as a general
    geometric-statistics library. JAX libraries with
    Riemannian-geometry support (Rieoptax, RiemannAX, GeoJAX) are surveyed in
    Appendix~\ref{app:jax}. $^\dagger$manify's 0.2.1 wheel on PyPI is incomplete and cannot be
    imported, we use git commit \texttt{18028ec}.}
    \label{tab:comparison}
    \begin{tabular}{@{}lccccc@{}}
      \toprule
      Feature & \textbf{hyperbolix} & \textbf{geoopt} & \textbf{HypLL} & \textbf{manify} & \textbf{Geomstats} \\
      \midrule
      Version                 & 1.3.0 & 0.5.1 & 0.1.1 & 0.2.1$^\dagger$ & 2.8.0 \\
      Framework               & \makecell{JAX/\\Flax NNX} & PyTorch & PyTorch & \makecell{PyTorch\\(on geoopt)} & \makecell{NumPy, Autograd,\\PyTorch} \\
      Hyp.\ models            & 3 (P, H, PV) & 2 (P, H) & 1 (P) & 2 (P, H) & 3 (P, H, half-sp.) \\
      Signed curvature        & \checkmark & \checkmark & $\times$ & \checkmark & $\times$ \\
      Product manifolds       & \checkmark & \checkmark & $\times$ & \checkmark & \checkmark \\
      Learnable curvature     & \checkmark & \checkmark & \checkmark & \checkmark & $\times$ \\
      Isometries              & \checkmark & \checkmark (P$\leftrightarrow$H) & $\times$ & \checkmark (P$\leftrightarrow$H) & \checkmark \\
      NN layers               & \checkmark (44) & $\times$ & \checkmark (11) & \checkmark (8) & $\times$ \\
      Optimizers              & Adam, SGD & \makecell{Adam, SGD, LS,\\sparse variants} & Adam, SGD & \makecell{Adan; Adam,\\SGD via geoopt} & Internal solvers \\
      Wrapped dist.           & P, H & P (sampling) & $\times$ & \makecell{H (P: sampling\\only)} & $\times$ \\
      Dim.\ reduction         & HoroPCA, CO-SNE & $\times$ & $\times$ & $\times$ & Tangent PCA, PGA \\
      License                 & MIT & Apache-2.0 & MIT & MIT & MIT \\
      \bottomrule
    \end{tabular}
  \end{table}

\section{Numerical Stability}
\label{sec:stability}

Distances in hyperbolic space grow exponentially with radius, so a fixed floating-point budget resolves exponentially less of the space with distance from the origin. On the Poincar\'e ball with curvature $c > 0$, the Riemannian metric is conformal to the Euclidean metric: for a tangent vector $v$, the conformal factor $\lambda(x) = 2/(1 - c\|x\|^2)$ rescales its Euclidean norm to its Riemannian norm at $x$, $\|v\|_x = \lambda(x)\|v\|_2$. It diverges as $\|x\|$ approaches the boundary radius $1/\sqrt{c}$. With float32 precision, spacing near $1.0$ is about $1.19 \times 10^{-7}$, so a boundary-adjacent quantity like $1 - c\|x\|^2$ loses most of its significant digits before any hyperbolic operation is executed.

\begin{figure}[t]
  \centering
  \includegraphics[width=\columnwidth]{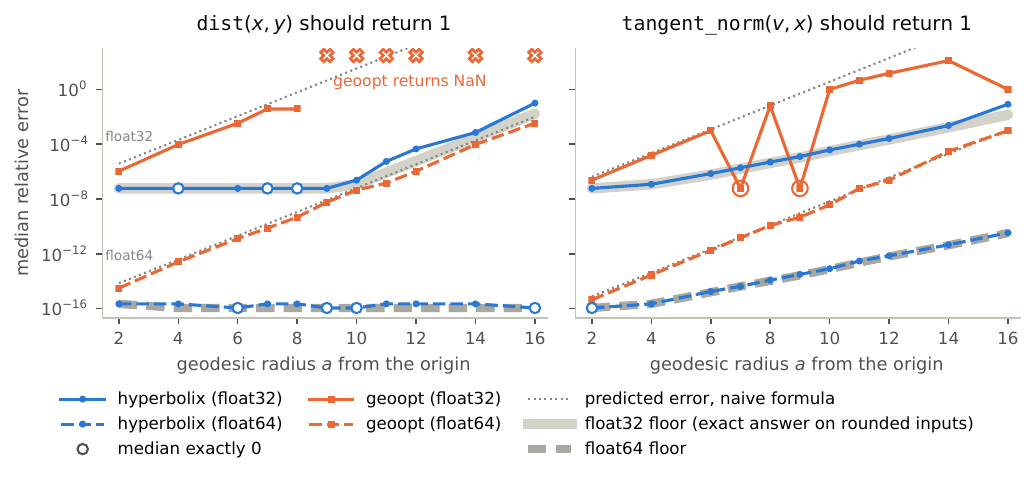}
  \caption{Median relative error of hyperboloid operations at curvature $c = 1$ in dimension $8$, as a function of the geodesic radius $a$ from the origin. We measure the distance between two points on one ray at unit separation, and the norm of a unit tangent vector at $45^\circ$ to the ray. For both float32 and float64, hyperbolix tracks the input-rounding error. On the float32 distance past $a \approx 10$, it stays within $8\times$ of it. For the tangent norm, geoopt's median is exactly $0$ at $a = 7$ and $9$ only because cancellation leaves it so few possible float32 results that more than half land exactly on $1$. The rest are far off.}
  \label{fig:cancellation_accuracy}
\end{figure}

The hyperboloid has its own distinct failure mode, caused by the Minkowski inner product $\langle x, y \rangle_L = -x_0 y_0 + \langle x_s, y_s \rangle$, where $x_0$ is a point's time coordinate and $x_s$ its spatial part. Almost every hyperboloid operation evaluates this inner product, including distance, tangent norm, parallel transport, and gyrovector addition. Its minus sign subtracts the product $x_0 y_0$ from the spatial inner product $\langle x_s, y_s \rangle$. For two nearby points at geodesic radius $a$, the hyperboloid constraint forces $\|x_s\| \approx x_0$, so $x_0 y_0$ and $\langle x_s, y_s \rangle$ both grow as $e^{2a}$~\citep{mishne2023numericalstability, vanDerKlis2026fgglnn}. These two large terms are nearly equal for nearby points. Because floating-point numbers store each term to a fixed number of digits, subtracting them erases the digits they share. The remaining absolute error is proportional to $e^{2a}$: for a large radius $a$, it can exceed the distance the computation is meant to measure. We call this \emph{ambient cancellation}. In float32, it exhausts the mantissa at $a \approx 9$ (float64: $a \approx 19$), even though the stored coordinates remain accurate to $a \approx 16$. Without mitigation, the error follows the predicted cancellation curve (Figure~\ref{fig:cancellation_accuracy}). geoopt returns NaN for distances at radii $a \approx 9$. An implementation that clamps the \texttt{acosh} input to its domain $[1, \infty)$ would return finite but wrong values after $a \approx 9$, instead.

\begin{figure}[t]
  \centering
  \includegraphics[width=\columnwidth]{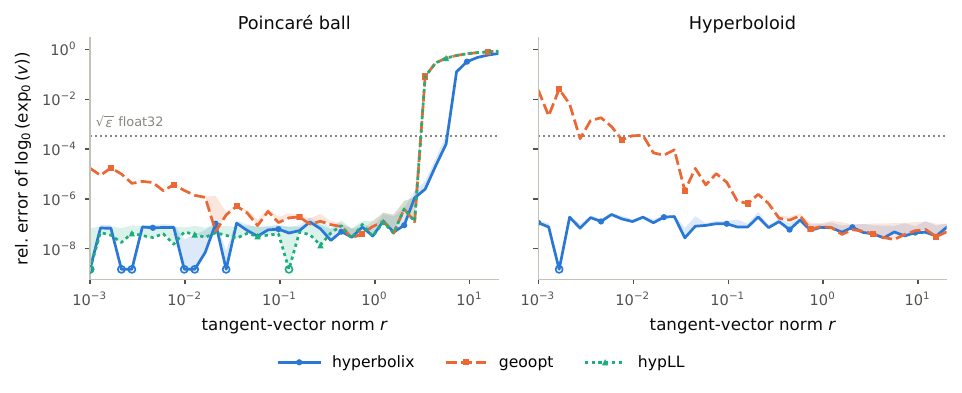}
  \caption{Expmap-logmap round trip stability of different libraries as a function of the tangent vector norm $r$. For the Poincar\'e ball, hyperbolix and HypLL are more stable than geoopt around the origin. On the hyperboloid, hyperbolix stays at the rounding level across the measured range.  geoopt shows elevated error near the origin. HypLL does not implement hyperboloid operations.}
  \label{fig:crosslib_stability}
\end{figure}

hyperbolix replaces those clamps with one rule: never form the canceling difference. Every Minkowski quadratic form is rewritten as a sum of non-negative terms. For the distance between two hyperboloid points $x$ and $y$ with spatial norms $r_x = \|x_s\|$, $r_y = \|y_s\|$, unit directions $\hat{x} = x_s / r_x$, $\hat{y} = y_s / r_y$, and radii $a_x = \operatorname{arsinh}(\sqrt{c}\, r_x)$, $a_y = \operatorname{arcosh}(\sqrt{c}\, y_0)$, this gives
\begin{equation}\label{eq:haversine}
  \sinh^2\!\bigl(\tfrac{\sqrt{c}\,d}{2}\bigr)
  \;=\;
  \sinh^2\!\bigl(\tfrac{a_x - a_y}{2}\bigr)
  \;+\;
  \tfrac{c}{4}\, r_x\, r_y\, \|\hat{x} - \hat{y}\|^2 \,,
\end{equation}
where $d$ is the geodesic distance. This yields the hyperbolic analog of the haversine identity, where the first term is the radial gap and the second the angular gap~\citep{Blasius2016EfficientEmbeddings, Celinska2024NumericalAspectsHyp}. Both are non-negative, and neither subtracts large, nearly equal quantities. The same decomposition of tangent vectors into radial and perpendicular components applies to the tangent metric, parallel transport, and gyrovector addition. In float32, the remaining error comes from storing the points, not from the formula: \textbf{hyperbolix's median distance stays within a factor of $8$ of the input-rounding error, the error that exact arithmetic on the float32-rounded points would still incur} (Figure~\ref{fig:cancellation_accuracy}). Beyond geodesic radius $a \approx 10$, the float32 distance error grows gradually rather than jumping to NaN: the median reaches $10^{-1}$ at $a = 16$ (Appendix~\ref{app:cancellation}). On the Poincar\'e ball, domain guards on \texttt{atanh} and \texttt{tanh} keep the gradients finite at the boundary; the cancellation problem does not arise there because the ball's arithmetic does not pass through a Minkowski inner product.  Figure~\ref{fig:crosslib_stability} compares the $\log_0(\exp_0(v))$ round-trip error across tangent-vector norms $r$ for geoopt~\citep{kochurov2020geoopt}, HypLL~\citep{vanspengler2023hypll}, and hyperbolix. On the Poincar\'e ball, hyperbolix and HypLL are comparable in accuracy, while geoopt exhibits elevated error near the origin. On the hyperboloid, hyperbolix stays at rounding error level for all tangent-vector norms $r$ from $10^{-3}$ to $20$. In contrast, geoopt's error is five orders of magnitude larger (median $2.3 \times 10^{-2}$ against $1.1 \times 10^{-7}$ at $r = 10^{-3}$) near the origin. It matches hyperbolix only for $r \gtrsim 1$. HypLL does not implement the hyperboloid.

Two conventions make stability checkable. First, tolerances are resolved through one function, $\texttt{default\_atol}(\mathrm{dtype}) = \sqrt{\texttt{finfo}(\mathrm{dtype}).\texttt{eps}}$, which gives $3.45 \times 10^{-4}$ in float32 and $1.49 \times 10^{-8}$ in float64. Every manifold-membership and tangent-space check uses it unless the caller overrides. On the Poincar\'e ball, membership tests the dimensionless residual $c\|x\|^2 - 1$, whereas on the hyperboloid it compares the stored time coordinate against the value derived from the spatial part. Both residuals are dimensionless, so the same tolerance applies uniformly across curvatures. Second, functions that admit several analytically equivalent formulations (e.g.\ distance computations) are selected by their \texttt{version\_idx} argument, static under \texttt{jax.jit}. These different versions exist for the Poincar\'e ball and the hyperboloid, where they trade speed against stability or gradient properties.

For operations anchored at the origin, the Poincar\'e ball loses accuracy before the other models. At $c = 1$, its median round-trip error exceeds $\sqrt{\varepsilon}$ between geodesic distances $11$ and $14.5$ in float32, and $19$ and $24$ in float64. In contrast, the hyperboloid and proper-velocity models remain at rounding error level in both dtypes to the end of our measured range, at a geodesic distance of $20$. In float32, the Poincar\'e ball's limit is set by hyperbolix's projection margin, which keeps points at least $\varepsilon^{0.75}$ inside the boundary and caps the ball at geodesic distance $12.6$. These radii scale as $1/\sqrt{c}$. Two-point hyperboloid operations lose accuracy at smaller radii than origin operations (Figure~\ref{fig:cancellation_accuracy}). Users who need more accuracy than float32 provides can run the manifold operations in float64 while keeping parameters stored in float32. The manifold's \texttt{dtype} sets the compute precision and each layer's \texttt{param\_dtype} the storage precision. \texttt{ProperVelocity} coordinates are unconstrained $\mathbb{R}^n$~\citep{chen2026pvnn} and have no boundary to drift toward. In hyperbolix, their two-point operations pass through the exact isometry to the hyperboloid, inheriting its cancellation-free arithmetic. The numbers behind both figures are in Appendix~\ref{app:stability_measure}.

\begin{figure}[t]
  \centering
  \includegraphics[width=\columnwidth]{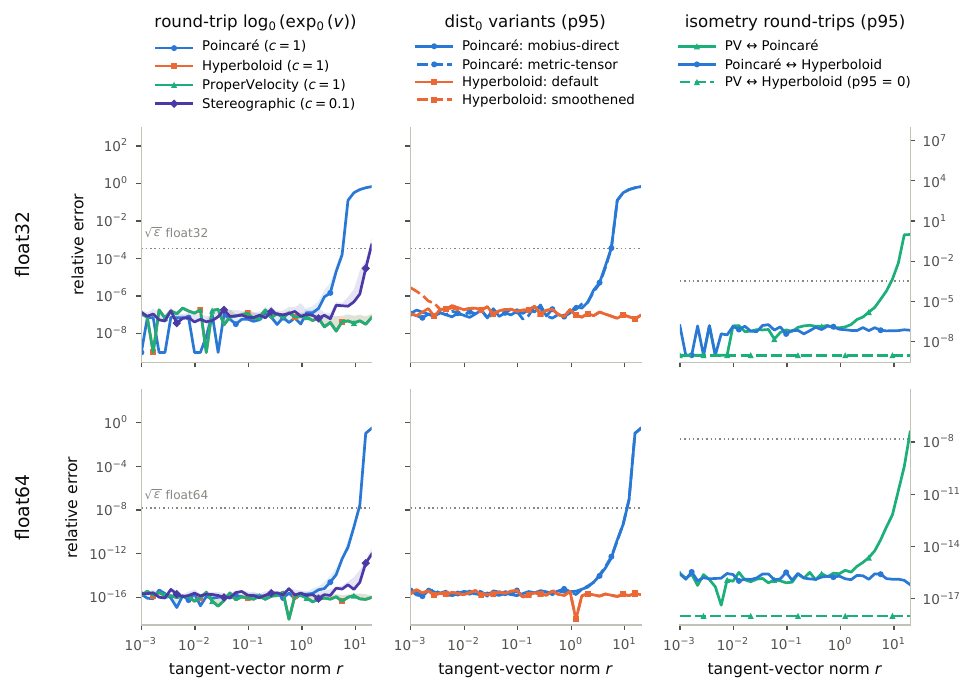}
  \caption{Relative error of hyperbolix operations in float32 (top) and float64 (bottom) as a function of the tangent-vector norm $r$, at curvature $c = 1$ in dimension $8$. \emph{Left:} origin round trip $\log_0(\exp_0(v))$. \emph{Center:} $\mathrm{dist}_0$ variants. \emph{Right:} isometry round-trip between manifold models. The $x$~axis is the tangent-vector norm. Geodesic distance is $2r$ on the ball and $\kappa$-stereographic model, $r$ elsewhere. The PV\,$\to$\,hyperboloid\,$\to$\,PV round trip is exactly zero by construction.}
  \label{fig:roundtrip_stability}
\end{figure}

\section{Correctness: Testing Against Independent Oracles}
\label{sec:correctness}

At curvature $c=1$, a hyperbolic operation whose closed form introduces a bug to its $\sqrt{c}$ scaling term is identical to the correct operation because every $\sqrt{c}$ factor equals $1$. At any other curvature, the error is silent: the operation still returns a point on the manifold, is still differentiable, and still trains. Shape checks, finiteness checks, and on-manifold checks all pass. Only a comparison against a known value catches the error, and such a comparison is worth exactly as much as the value it compares against. Every checked operation in hyperbolix is therefore tested against an \emph{independent oracle}: a hand-derived closed form, a NumPy/SciPy transcription of the defining expression from the source paper, or a finite-difference estimate. No correctness assertion compares one hyperbolix call against another or against a second library~\citep{kochurov2020geoopt, vanspengler2023hypll} because a self-consistent implementation passes its own cross-checks. Consequently, two libraries derived from the same reference code would inherit the same mistakes. Comparisons between hyperbolix implementations serve only as \emph{consistency checks}: \texttt{jit} against eager execution, tangent-space against manifold input, and expmap-logmap round-trips. Our test suite comprises 1,038 test functions across 55 files, parametrized over float32 and float64. Appendix~\ref{app:tests} details the tolerance conventions, the shared layer contract, and these consistency checks. A transcription oracle, however, verifies correctness versus the source paper, not the paper's correctness.

An illustrative example highlighting the benefits of this design is \texttt{Hyperboloid.log\_radius\_concat}, which uses the digamma function $\psi$ to rescale the spatial part of each of $N$ blocks of dimension $d$ so that the expected log spatial radius is preserved. Swapping the two digamma arguments turns the intended shrinkage $\exp(\tfrac{1}{2}(\psi(d/2) - \psi(Nd/2))) \approx 1/\sqrt{N}$ into a concatenated radius that is $\mathcal{O}(N)$ too large, where $N$ is the number of blocks and $d$ the block dimension. The concatenation would inflate the radius that the rescaling exists to preserve. Generic on-manifold tests cannot catch this bug: the output is still a valid manifold point. The same swap is present in the ``Intrinsic Lorentz Neural Network'' paper and reference implementation~\citep{shi2026ilnn}. A test based on that implementation would have reproduced the error, whereas the oracle caught it due to it being derived from the identity $\mathbb{E}[\log \|v\|] = \tfrac{1}{2}(\psi(k/2) + \log 2)$ for $\|v\|^2 \sim \chi^2_k$, where  $\chi^2_k$ is the Chi-squared distribution. Table~\ref{tab:oracle} reports per-operation deviations between each hyperbolix call and its oracle over the standard sampling grids.

\section{Performance}
\label{sec:performance}

\begin{figure}[t]
  \centering
  \includegraphics[width=\columnwidth]{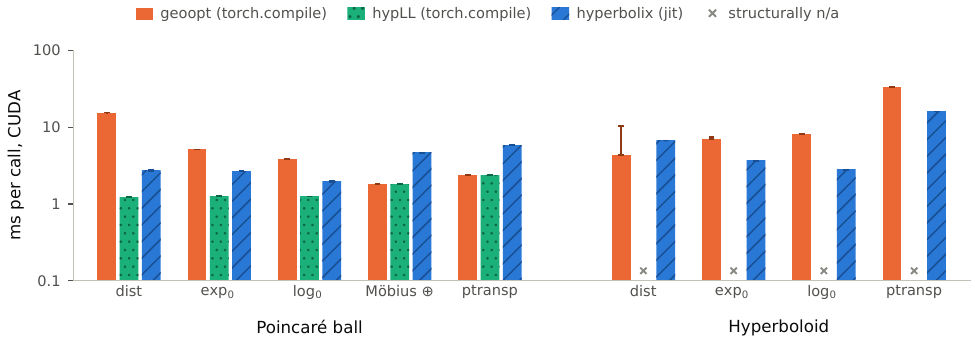}
  \caption{Wall-clock time per call for matched primitives on CUDA (H100, float32, batch $10^7$, five launches at seed~0). geoopt~\citep{kochurov2020geoopt} and HypLL~\citep{vanspengler2023hypll} use \texttt{torch.compile}, while hyperbolix uses \texttt{jax.jit} with \texttt{vmap}. Brackets span the five launches and bound machine and compilation state, not sampling variation. HypLL implements only the Poincar\'e ball. The $y$-axis is logarithmic with a 0.1\,ms floor.}
  \label{fig:cuda_speed}
\end{figure}

In contrast to Euclidean layers, hyperbolic layers incur additional pointwise overhead due to manifold operations. A M\"obius linear map or a Lorentz distance evaluates transcendentals elementwise, then threads the result through a chain of pointwise steps (norms, clamps, conformal rescalings, re-projections) whose only purpose is to return a valid manifold point. Each of these steps is memory-bound: it moves more data than the arithmetic it performs would require. Under eager execution, each step also runs as its own kernel, so every step pays a separate kernel launch and a separate round trip to memory. \texttt{jax.jit} fuses many of these steps into fewer kernels, keeps intermediates in registers where possible, and removes the Python dispatch that otherwise dominates at small batch sizes~\citep{bradbury2018jax}. Because hyperbolix defines each manifold operation on a single point (Section~\ref{sec:design}), \texttt{jax.vmap} covers every batch shape from that one definition without hand-written broadcasting.

Figure~\ref{fig:cuda_speed} shows wall-clock time per call for matched primitives (distance,
$\exp_0$, $\log_0$, M\"obius addition, parallel transport) in float32 using CUDA on a batch of size $10^7$. We compare geoopt~
\citep{kochurov2020geoopt}, HypLL~\citep{vanspengler2023hypll}, and hyperbolix. The large batch size is required to amortize dispatch overhead. geoopt and HypLL use \texttt{torch.compile} to match \texttt{jax.jit}, and hyperbolix operations are batched with \texttt{vmap}. The results show that HypLL is the fastest for the only manifold it implements, the Poincar\'e ball. geoopt and hyperbolix are broadly comparable in speed, with the winner dependent on the exact function being tested.

\section{Limitations}
\label{sec:limitations}

The measurements of Sections~\ref{sec:correctness}--\ref{sec:performance} have a narrow scope. The
oracle table (Table~\ref{tab:oracle}), the cross-library accuracy table
(Table~\ref{tab:crosslib_accuracy}), the stability figures (Figures~\ref{fig:crosslib_stability}
and~\ref{fig:roundtrip_stability}), and the speed comparison (Figure~\ref{fig:cuda_speed};
Tables~\ref{tab:speed}--\ref{tab:speed_cpu_default}) are single-seed sweeps: every speed launch uses
seed 0, so the brackets bound launch-to-launch machine and compilation state, and not sampling
variation. We report no training-quality benchmarks: nothing here shows hyperbolix reaching a published accuracy on a published task.

Downstream evidence is thin. The library was developed and hardened within one research project; the
two vector-quantization bottlenecks were ported from it. Wider use is what would expose interfaces
that only appear general.

Two limits are structural. The float32 range ceiling of Section~\ref{sec:stability} follows from the
exponential growth of the hyperbolic metric. For the hyperboloid, the cancellation-free formulas of Section~\ref{sec:stability} push the failure point outward without removing it. For the Poincar\'e ball at $c = 1$, past a geodesic radius of about $14$ float64 remains the only remedy, and past about $24$ it fails in turn. Coverage is also uneven across manifolds: the three attention layers and both positional encodings exist only on the hyperboloid, and both vector-quantization bottlenecks only on the Poincar\'e ball (Appendix~\ref{app:layers}).

\section{Conclusion}\label{sec:conclusion}
We present hyperbolix, a hyperbolic deep learning library for JAX and Flax NNX. To our knowledge, it is the first general-purpose hyperbolic deep learning library in JAX. hyperbolix comprises six manifolds using a common interface, layer families covering linear layers, convolutions, attention, normalization, positional encoding, regression, and vector quantization, Riemannian optimizers written as optax transformations. Its API is written to enable \texttt{jit}, \texttt{vmap}, and \texttt{scan}. The correctness of hyperbolix is based on the independent-oracle test suite of Section~\ref{sec:correctness}. The tests compare every checked operation against a hand-derived closed form, a transcription of the source paper, or a finite difference rather than against another library call. \emph{hyperbolix's cancellation-free hyperboloid arithmetic keeps the float32 distance error within $8\times$ of the input-rounding error (Section~\ref{sec:stability}).} We report comprehensive measurements, and make hyperbolix available under the MIT license at \url{https://github.com/timoklein/hyperbolix}. Contributions are welcome.

\bibliographystyle{tmlr}
\bibliography{refs}

\appendix
\makeatletter
\newenvironment{apptable}
  {\par\addvspace{\intextsep}\noindent\begin{minipage}{\linewidth}%
   \def\@captype{table}\centering}
  {\end{minipage}\par\addvspace{\intextsep}}
\newenvironment{appfigure}
  {\par\addvspace{\intextsep}\noindent\begin{minipage}{\linewidth}%
   \def\@captype{figure}\centering}
  {\end{minipage}\par\addvspace{\intextsep}}
\makeatother

\section{Layer Inventory}
\label{app:layers}

This appendix enumerates all 76 public exports of \texttt{hyperbolix.nn\_layers} (the \texttt{\_\_all\_\_} list of \texttt{hyperbolix/nn\_layers/\_\_init\_\_.py}, hyperbolix v1.3.0), grouped into ten families. ``Type'' distinguishes an \texttt{nnx.Module} (stateful, owns parameters/buffers) from a plain JAX function. Manifold abbreviations: \textsc{Hyp} = Hyperboloid (Lorentz) model, \textsc{Poin} = Poincar\'e ball, \textsc{PV} = Proper Velocity.

\subsection{Linear}

\begin{apptable}
\centering
\begin{tabular}{@{}llll@{}}
\toprule
Export & Manifold(s) & Type & Source \\
\midrule
\texttt{HTCLinear} & Hyp & Module & \citep{yang2025hypformer} \\
\texttt{FGGLinear} & Hyp & Module & \citep{vanDerKlis2026fgglnn} \\
\texttt{HypLinearHyperboloidPLFC} & Hyp & Module & \citep{shi2026ilnn} \\
\texttt{HypLinearHyperboloidFHCNN} & Hyp & Module & \citep{bdeir2024fhcnn} \\
\texttt{HypLinearHyperboloidFHNN} & Hyp & Module & \citep{chen2022fhnn} \\
\texttt{HypLinearHyperboloidBusemann} & Hyp & Module & \citep{chen2026busemann_networks} \\
\texttt{HypLinearPoincarePP} & Poin & Module & \citep{shimizu2021hnnpp} \\
\texttt{HypLinearPoincare} & Poin & Module & \citep{ganea2018hnn} \\
\texttt{HypLinearPoincareBusemann} & Poin & Module & \citep{chen2026busemann_networks} \\
\texttt{HypLinearPV} & PV & Module & \citep{chen2026pvnn} \\
\texttt{busemann\_fc\_poincare\_output} & Poin & function & \citep{chen2026busemann_networks} \\
\bottomrule
\end{tabular}
\caption{Linear layers (11 exports: ten layer classes and one function). \texttt{HypLinearPoincare} and \texttt{HypRegressionPoincare} (below) are the only layers in the library whose weights are manifold-valued and require a Riemannian optimizer; every other layer in this table trains with plain \texttt{optax.adam}.}
\end{apptable}

\subsection{Convolution}

\begin{apptable}
\centering
\begin{tabular}{@{}llll@{}}
\toprule
Export & Manifold(s) & Type & Source \\
\midrule
\texttt{HypConv2DHyperboloid} & Hyp & Module & \citep{bdeir2024fhcnn} \\
\texttt{HypConv2DHyperboloidFHNN} & Hyp & Module & \citep{chen2022fhnn} \\
\texttt{HypConv2DHyperboloidILNN} & Hyp & Module & \citep{shi2026ilnn} \\
\texttt{FGGConv2D} & Hyp & Module & \citep{vanDerKlis2026fgglnn} \\
\texttt{LorentzConv2D} & Hyp & Module & \citep{he2025lresnet} \\
\texttt{HypConv2DPoincare} & Poin & Module & \citep{shimizu2021hnnpp,vanspengler2023resnet} \\
\texttt{HypConv2DPV} & PV & Module & \citep{chen2026pvnn} \\
\bottomrule
\end{tabular}
\caption{Convolutional layers (7 exports). \texttt{LorentzConv2D} is kept for reproduction of the LResNet convolutional stack; it is HRC-based (Euclidean conv on the spatial components, time reconstructed from the hyperboloid constraint) and is superseded by \texttt{HypConv2DHyperboloid} for new work.}
\end{apptable}

\subsection{Attention}

\begin{apptable}
\centering
\begin{tabular}{@{}llll@{}}
\toprule
Export & Manifold(s) & Type & Source \\
\midrule
\texttt{HyperbolicLinearAttention} & Hyp & Module & \citep{yang2025hypformer} \\
\texttt{HyperbolicSoftmaxAttention} & Hyp & Module & \citep{yang2025hypformer} \\
\texttt{HyperbolicFullAttention} & Hyp & Module & \citep{yang2025hypformer} \\
\texttt{focus\_transform} & Hyp & function & \citep{yang2025hypformer} \\
\bottomrule
\end{tabular}
\caption{Attention layers (4 exports: three attention layers and one shared function). \texttt{HyperbolicLinearAttention} is $O(N)$ (kernel-trick linear attention with the focus function of \texttt{focus\_transform}); \texttt{HyperbolicSoftmaxAttention} and \texttt{HyperbolicFullAttention} are $O(N^2)$, operating in the spatial domain and on full Lorentzian inner products respectively.}
\end{apptable}

\subsection{Normalization \& Dropout}

\begin{apptable}
\centering
\begin{tabular}{@{}llll@{}}
\toprule
Export & Manifold(s) & Type & Source \\
\midrule
\texttt{HyperboloidGyroBatchNorm} & Hyp & Module & \citep{chen2025gyrobn} \\
\texttt{HyperboloidGyroRMSNorm} & Hyp & Module & \citep{chen2025gyrobn} \\
\texttt{PoincareGyroRMSNorm} & Poin & Module & \citep{chen2025gyrobn} \\
\texttt{ProperVelocityGyroBatchNorm} & PV & Module & \citep{chen2025gyrobn} \\
\texttt{ProperVelocityGyroRMSNorm} & PV & Module & \citep{chen2025gyrobn} \\
\texttt{PoincareBatchNorm2D} & Poin & Module & \citep{vanspengler2023resnet} \\
\texttt{HRCLayerNorm} & Hyp & Module & \citep{yang2025hypformer} \\
\texttt{HRCRMSNorm} & Hyp & Module & \citep{yang2025hypformer} \\
\texttt{HRCBatchNorm} & Hyp & Module & \citep{yang2025hypformer} \\
\texttt{HRCDropout} & Hyp & Module & \citep{yang2025hypformer} \\
\texttt{FGGMeanOnlyBatchNorm} & Hyp & Module & \citep{vanDerKlis2026fgglnn} \\
\texttt{frechet\_variance} & any Manifold & function & \citep{vanspengler2023resnet} \\
\bottomrule
\end{tabular}
\caption{Normalization and dropout layers (12 exports: ten normalization layers, one dropout layer, and one shared function). \texttt{*GyroBatchNorm}/\texttt{*GyroRMSNorm} are intrinsic (gyrovector) normalizers operating directly on manifold points; \texttt{HRC*} normalizers wrap a Euclidean \texttt{flax.nnx} layer on the spatial components via the Hypformer HRC pattern. \texttt{frechet\_variance} is manifold-generic (used by both \texttt{PoincareBatchNorm2D} and the Gyro-BatchNorm family) and takes any object satisfying the \texttt{Manifold} protocol.}
\end{apptable}

\subsection{Residual, Pooling \& Primitives}

\begin{apptable}
\centering
\begin{tabular}{@{}llll@{}}
\toprule
Export & Manifold(s) & Type & Source \\
\midrule
\texttt{LorentzResidual} & Hyp & Module & \citep{he2025lresnet} \\
\texttt{lorentz\_residual} & Hyp & function & \citep{he2025lresnet} \\
\texttt{lorentz\_scale} & Hyp & function & \citep{he2025lresnet} \\
\texttt{lorentz\_midpoint} & Hyp & function & \citep{he2025hope_helm} \\
\texttt{hyp\_avg\_pool2d} & Hyp & function & \citep{yang2025hypformer} \\
\texttt{hyp\_flatten2d} & Hyp & function & \citep{shi2026ilnn} \\
\texttt{hrc} & Hyp & function & \citep{yang2025hypformer} \\
\texttt{htc} & Hyp & function & \citep{yang2025hypformer} \\
\texttt{extract\_patches} & Hyp & function & N/A \\
\texttt{hcat\_ambient\_dim} & Hyp & function & N/A \\
\texttt{build\_spacelike\_V} & Hyp & function & \citep{vanDerKlis2026fgglnn} \\
\texttt{spatial\_to\_hyperboloid} & Hyp & function & \citep{yang2025hypformer} \\
\texttt{sinh\_lift\_to\_hyperboloid} & Hyp & function & \citep{shi2026ilnn} \\
\texttt{poincare\_midpoint} & Poin & function & \citep{ungar2009gyrovector, vanspengler2023resnet} \\
\texttt{poincare\_weighted\_midpoint} & Poin & function & \citep{Bu2025GGBall} \\
\bottomrule
\end{tabular}
\caption{Residual connections, pooling, and shared low-level primitives (15 exports). \texttt{extract\_patches} and \texttt{hcat\_ambient\_dim} are shared infrastructure behind every hyperboloid conv layer (im2col-style patch extraction and the HCat/LogCat output-dimension formula) with no single-paper attribution. \texttt{lorentz\_midpoint} generalizes \texttt{lorentz\_residual} (two points) to a weighted midpoint over $M$ points via the closed-form Lorentz centroid of \citep{he2025hope_helm}; it backs \texttt{HyperbolicFullAttention}'s aggregation and \texttt{HyperboloidGyroBatchNorm}'s batch-mean estimator. \texttt{poincare\_midpoint}/\texttt{poincare\_weighted\_midpoint} implement the Einstein/gyromidpoint used by \texttt{PoincareBatchNorm2D} and the Poincar\'e VQ codebook EMA update~\citep{Bu2025GGBall}.}
\end{apptable}

\subsection{Positional Encoding}

\begin{apptable}
\centering
\begin{tabular}{@{}llll@{}}
\toprule
Export & Manifold(s) & Type & Source \\
\midrule
\texttt{HypformerPositionalEncoding} & Hyp & Module & \citep{yang2025hypformer} \\
\texttt{HyperbolicRoPE} & Hyp & Module & \citep{he2025hope_helm} \\
\texttt{hope} & Hyp & function & \citep{he2025hope_helm} \\
\bottomrule
\end{tabular}
\caption{Positional encoding (3 exports). \texttt{HypformerPositionalEncoding} is a learnable relative encoding (\texttt{HTCLinear} + fixed-weight Lorentzian residual); \texttt{hope}/\texttt{HyperbolicRoPE} is the deterministic, rotation-based Hyperbolic Rotary Positional Encoding.}
\end{apptable}

\subsection{Regression / MLR}

\begin{apptable}
\centering
\begin{tabular}{@{}llll@{}}
\toprule
Export & Manifold(s) & Type & Source \\
\midrule
\texttt{HypRegressionHyperboloid} & Hyp & Module & \citep{bdeir2024fhcnn} \\
\texttt{FGGLorentzMLR} & Hyp & Module & \citep{vanDerKlis2026fgglnn} \\
\texttt{HypRegressionHyperboloidBusemann} & Hyp & Module & \citep{chen2026busemann_networks} \\
\texttt{HypRegressionPoincarePP} & Poin & Module & \citep{shimizu2021hnnpp} \\
\texttt{HypRegressionPoincare} & Poin & Module & \citep{ganea2018hnn} \\
\texttt{HypRegressionPoincareBusemann} & Poin & Module & \citep{chen2026busemann_networks} \\
\texttt{HypRegressionPV} & PV & Module & \citep{chen2026pvnn} \\
\bottomrule
\end{tabular}
\caption{Multinomial logistic regression / classification heads (7 exports). All but \texttt{HypRegressionPoincare} return Euclidean logits from a point-to-hyperplane (or, for the Busemann pair, point-to-horosphere) score and train with plain \texttt{optax.adam}.}
\end{apptable}

\subsection{Vector Quantization}

\begin{apptable}
\centering
\begin{tabular}{@{}llll@{}}
\toprule
Export & Manifold(s) & Type & Source \\
\midrule
\texttt{HypVQEmbeddingPoincare} & Poin & Module & \citep{chen2025hvqvae} \\
\texttt{HypVQMLRPoincare} & Poin & Module & \citep{goswami2025hypervq} \\
\texttt{PoincareVQOutput} & Poin & NamedTuple & \citep{chen2025hvqvae,goswami2025hypervq} \\
\bottomrule
\end{tabular}
\caption{Poincar\'e-ball vector quantization (3 exports). \texttt{HypVQEmbeddingPoincare} is an explicit EMA codebook (buffer, not a trainable parameter) with a copy-gradient straight-through estimator; \texttt{HypVQMLRPoincare} is an implicit codebook realized as the rows of a Poincar\'e MLR, selected via Gumbel-Softmax quantization-as-classification and trainable end to end with Euclidean Adam. \texttt{PoincareVQOutput} is the shared \texttt{NamedTuple} return type of both.}
\end{apptable}

\subsection{Activations}

\begin{apptable}
\centering
\begin{tabular}{@{}llll@{}}
\toprule
Export & Manifold(s) & Type & Source \\
\midrule
\texttt{hrc\_relu} & Hyp & function & \citep{yang2025hypformer} \\
\texttt{hrc\_leaky\_relu} & Hyp & function & \citep{yang2025hypformer} \\
\texttt{hrc\_tanh} & Hyp & function & \citep{yang2025hypformer} \\
\texttt{hrc\_swish} & Hyp & function & \citep{yang2025hypformer} \\
\texttt{hrc\_gelu} & Hyp & function & \citep{yang2025hypformer} \\
\texttt{hyp\_relu} & Hyp & function & \citep{bdeir2024fhcnn} \\
\texttt{hyp\_leaky\_relu} & Hyp & function & \citep{bdeir2024fhcnn} \\
\texttt{hyp\_tanh} & Hyp & function & \citep{bdeir2024fhcnn} \\
\texttt{hyp\_swish} & Hyp & function & \citep{bdeir2024fhcnn} \\
\texttt{hyp\_gelu} & Hyp & function & \citep{bdeir2024fhcnn} \\
\texttt{poincare\_relu} & Poin & function & \citep{vanspengler2023resnet} \\
\texttt{poincare\_leaky\_relu} & Poin & function & \citep{vanspengler2023resnet} \\
\texttt{poincare\_tanh} & Poin & function & \citep{vanspengler2023resnet} \\
\bottomrule
\end{tabular}
\caption{Activation functions (13 exports). \texttt{hrc\_*} are curvature-changing ($c_{\mathrm{in}} \neq c_{\mathrm{out}}$) convenience wrappers around \texttt{hrc}; \texttt{hyp\_*} are the curvature-preserving special case ($c_{\mathrm{in}}=c_{\mathrm{out}}$), documented against the fully-hyperbolic-CNN reference. \texttt{poincare\_*} implement $f_{\mathcal{P}} = \exp_0^c \circ f \circ \log_0^c$ (activation applied in the tangent space at the origin).}
\end{apptable}

\subsection{Hybrid / Interface}

\begin{apptable}
\centering
\begin{tabular}{@{}llll@{}}
\toprule
Export & Manifold(s) & Type & Source \\
\midrule
\texttt{HyperPPFeatureScaling} & Euclidean $\to$ Poin/Hyp & Module & \citep{klein2026hyperpp} \\
\bottomrule
\end{tabular}
\caption{Euclidean-to-hyperbolic boundary layer (1 export). \texttt{HyperPPFeatureScaling} is not itself on-manifold: it applies parameter-free RMSNorm, a Lipschitz activation, and $1/\sqrt{d}$ dimension scaling to Euclidean features immediately before \texttt{expmap\_0}, optionally followed by a learned sigmoid-bounded rescale $\rho_{\max}=\operatorname{atanh}(\alpha)/\sqrt{c}$ (Hyper++, \citep{klein2026hyperpp}, Sec.\ 3.2) that keeps the first on-manifold point strictly inside the ball/off the boundary.}
\end{apptable}

\section{Implemented Methods}
\label{app:methods}

Table~\ref{tab:methods} maps published methods to the hyperbolix components that implement them, gathered from module- and class-level docstrings across the package (not only \texttt{nn\_layers}).

\begin{apptable}
\centering
\footnotesize
\begin{tabular}{@{}p{3.4cm}p{8.4cm}l@{}}
\toprule
Method & hyperbolix components & Reference \\
\midrule
Hyperbolic Neural Networks (HNN) & \texttt{HypLinearPoincare}, \texttt{HypRegressionPoincare}, \texttt{ManifoldParam} + \texttt{riemannian\_adam} & \citep{ganea2018hnn} \\
HNN++ (point-to-hyperplane FC/MLR) & \texttt{HypLinearPoincarePP}, \texttt{HypRegressionPoincarePP}, \texttt{HypConv2DPoincare} & \citep{shimizu2021hnnpp} \\
Poincar\'e ResNet & \texttt{HypConv2DPoincare}, \texttt{PoincareBatchNorm2D}, \texttt{poincare\_relu}, \texttt{HyperPPFeatureScaling} & \citep{vanspengler2023resnet} \\
Fully Hyperbolic Neural Networks (FHNN) & \texttt{HypLinearHyperboloidFHNN}, \texttt{HypConv2DHyperboloidFHNN} & \citep{chen2022fhnn} \\
Fully hyperbolic CNN (HCat) & \texttt{HypConv2DHyperboloid}, \texttt{HypLinearHyperboloidFHCNN}, \texttt{hyp\_*} activations, \texttt{HypRegressionHyperboloid} & \citep{bdeir2024fhcnn} \\
Hypformer HRC (Hyperbolic Regularization Component) & \texttt{hrc}, \texttt{hrc\_*} activations, \texttt{HRCLayerNorm}, \texttt{HRCRMSNorm}, \texttt{HRCBatchNorm}, \texttt{HRCDropout}, \texttt{LorentzConv2D} & \citep{yang2025hypformer} \\
Hypformer HTC (Hyperbolic Transformation Component) & \texttt{htc}, \texttt{HTCLinear} & \citep{yang2025hypformer} \\
Hypformer attention & \texttt{HyperbolicLinearAttention}, \texttt{HyperbolicSoftmaxAttention}, \texttt{HyperbolicFullAttention}, \texttt{focus\_transform} & \citep{yang2025hypformer} \\
Hypformer positional encoding & \texttt{HypformerPositionalEncoding} & \citep{yang2025hypformer} \\
Fast \& Geometrically Grounded LNN (FGG-LNN) & \texttt{FGGLinear}, \texttt{FGGConv2D}, \texttt{FGGLorentzMLR}, \texttt{FGGMeanOnlyBatchNorm}, \texttt{build\_spacelike\_V} & \citep{vanDerKlis2026fgglnn} \\
Intrinsic Lorentz NN (ILNN), point-to-hyperplane FC (PLFC) & \texttt{HypLinearHyperboloidPLFC}, \texttt{sinh\_lift\_to\_hyperboloid} & \citep{shi2026ilnn} \\
ILNN LogCat convolution & \texttt{HypConv2DHyperboloidILNN}, \texttt{hyp\_flatten2d}, \texttt{Hyperboloid.log\_radius\_concat} & \citep{shi2026ilnn} \\
Intrinsic Lorentz gyrovector operations & \texttt{Hyperboloid.addition}, \texttt{Hyperboloid.scalar\_mul} & \citep{shi2026ilnn} \\
Proper Velocity Neural Networks (PVNN) & \texttt{HypLinearPV}, \texttt{HypConv2DPV}, \texttt{HypRegressionPV}, \texttt{ProperVelocity} manifold & \citep{chen2026pvnn} \\
Lorentzian ResNet (LResNet) & \texttt{LorentzResidual}, \texttt{lorentz\_residual}, \texttt{lorentz\_scale}, \texttt{LorentzConv2D} & \citep{he2025lresnet} \\
Hyperbolic Busemann Networks & \texttt{HypLinearHyperboloidBusemann}, \texttt{HypLinearPoincareBusemann}, \texttt{HypRegressionHyperboloidBusemann}, \texttt{HypRegressionPoincareBusemann} & \citep{chen2026busemann_networks} \\
Gyrogroup Batch Normalization (GyroBN) & \texttt{HyperboloidGyroBatchNorm}, \texttt{ProperVelocityGyroBatchNorm}, \texttt{*GyroRMSNorm} family & \citep{chen2025gyrobn} \\
Hyperbolic Rotary Positional Encoding (HOPE / HELM) & \texttt{hope}, \texttt{HyperbolicRoPE}, \texttt{lorentz\_midpoint} & \citep{he2025hope_helm} \\
Hyperbolic VQ-VAE, EMA codebook (HVQ-VAE) & \texttt{HypVQEmbeddingPoincare}, \texttt{poincare\_weighted\_midpoint} & \citep{chen2025hvqvae} \\
Hyperbolic VQ-VAE, quantization-as-classification (HyperVQ) & \texttt{HypVQMLRPoincare} & \citep{goswami2025hypervq} \\
$\kappa$-Stereographic model & \texttt{Stereographic} manifold class & \citep{bachmann2020kappa_stereographic} \\
HoroPCA & \texttt{hyperbolix.decomposition.horopca} & \citep{chami2021horopca} \\
CO-SNE & \texttt{hyperbolix.decomposition.cosne} & \citep{guo2022cosne} \\
Fr\'echet / Karcher mean & \texttt{hyperbolix.decomposition.frechet.frechet\_mean} & \citep{lou2020frechet_mean} \\
Wrapped normal distribution & \texttt{wrapped\_normal\_hyperboloid}, \texttt{wrapped\_normal\_poincare} & \citep{Nagano2019wrapped_normal} \\
Riemannian Adam & \texttt{hyperbolix.optim.riemannian\_adam}, \texttt{ManifoldParam} & \citep{becigneul2018riemannian} \\
Riemannian SGD & \texttt{hyperbolix.optim.riemannian\_sgd} & \citep{bonnabel2013rsgd} \\
Product (mixed-curvature) manifolds & \texttt{ProductManifold}, per-factor \texttt{LearnableCurvature} & \citep{gu2019mixed} \\
Poincar\'e embeddings (smoothened distance) & \texttt{Hyperboloid.dist}/\texttt{dist\_0}, \texttt{VERSION\_SMOOTHENED} & \citep{nickel2017poincareembeddings} \\
Gyrovector formalism & \texttt{Poincare}/\texttt{Hyperboloid}/\texttt{ProperVelocity}\ \texttt{addition}, \texttt{scalar\_mul} & \citep{ungar2009gyrovector} \\
Delta-hyperbolicity / Gromov product & \texttt{hyperbolix.utils.helpers.compute\_hyperbolic\_delta} & \citep{khrulkov2020gromov_hyperbolicity} \\
\bottomrule
\end{tabular}
\caption{Implemented methods and the hyperbolix components realizing them (31 rows).}
\label{tab:methods}
\end{apptable}

\section{The JAX Ecosystem}
\label{app:jax}

Table~\ref{tab:jax_ecosystem} documents the three JAX libraries with a focus similar to hyperbolix that provide some Riemannian or hyperbolic functionality, verified against their default-branch sources on 2026-08-03. None offers a general-purpose hyperbolic layer library. Rieoptax contains nine undocumented Poincar\'e/Lorentz layer classes (no convolutions) and has been inactive since October 2023. RiemannAX's hyperbolic manifolds exist only in unreleased development code. GeoJAX, created in July 2026 and currently under active development, implements both the Poincaré ball and hyperboloid models with a broad suite of Riemannian optimization methods. However, it contains no neural-network layers.

\begin{apptable}
\centering
\small
\begin{tabular}{@{}lp{3.1cm}p{3.3cm}p{3.6cm}@{}}
\toprule
 & Rieoptax & RiemannAX & GeoJAX \\
\midrule
Stated focus & Riemannian optim., diff.\ privacy & Riemannian optim. & Riemannian geometry, optim., classical ML \\
Hyperbolic manifolds & P, H & P, H (unreleased) & P, H \\
Hyperbolic NN layers & 9 undocumented classes, no conv & No (one manifold-agnostic linear) & No \\
Riemannian optimizers & SGD (Adam stubbed) & SGD, Adam, momentum & First- and second-order suite \\
Last activity (as of 2026-08) & Oct.\ 2023 & Mar.\ 2026 & Active \\
License & MIT (metadata only) & Apache-2.0 & MIT \\
\bottomrule
\end{tabular}
\caption{JAX libraries with Riemannian or hyperbolic functionality, from their default-branch sources as of 2026-08-03. P: Poincar\'e ball, H: hyperboloid. Rieoptax lists MIT only in its package metadata; the repository ships no license file.}
\label{tab:jax_ecosystem}
\end{apptable}

\section{The Manifold Protocol}
\label{app:protocol}

\texttt{hyperbolix.manifolds.protocol.Manifold} is a \texttt{typing.Protocol} decorated \texttt{@runtime\_checkable}: no concrete manifold class inherits from it, and \texttt{isinstance(x, Manifold)} succeeds for any object that exposes the right method signatures. This structural typing is what lets \texttt{ProductManifold} satisfy the protocol without a shared base class — it is a plain composition of factor manifolds that happens to expose the same 19-method surface (its \texttt{c} argument is a per-factor \emph{sequence} rather than a scalar, which the protocol's \texttt{Curvature = ScalarCurvature | Sequence[ScalarCurvature]} union accommodates). \texttt{Poincare}, \texttt{Hyperboloid}, \texttt{ProperVelocity}, \texttt{Euclidean}, and \texttt{Stereographic} all conform the same way.

Table~\ref{tab:protocol} lists the 19 required methods (the \texttt{dtype} field is an attribute, not a method, and is excluded from the count) with one-line semantics.

\begin{apptable}
\centering
\small
\begin{tabular}{@{}p{5.1cm}p{11.0cm}@{}}
\toprule
Method & Semantics \\
\midrule
\texttt{\_cast(x)} & Cast an input array to the manifold's compute dtype. \\
\texttt{proj(x, c)} & Project a point back onto the manifold's defining constraint. \\
\texttt{dist(x, y, c)} & Geodesic distance between two manifold points. \\
\texttt{dist\_0(x, c)} & Geodesic distance from a point to the manifold origin. \\
\texttt{addition(x, y, c)} & Gyrovector (M\"obius / Lorentz) addition of two manifold points. \\
\texttt{scalar\_mul(r, x, c)} & Gyrovector scalar multiplication; scales the geodesic radius by $r$. \\
\texttt{expmap(v, x, c)} & Riemannian exponential map: move from $x$ along tangent $v$ to a manifold point. \\
\texttt{expmap\_0(v, c)} & Exponential map at the origin ($v$ assumed tangent there). \\
\texttt{logmap(y, x, c)} & Riemannian logarithmic map (inverse of \texttt{expmap}) from $x$ to $y$. \\
\texttt{logmap\_0(y, c)} & Logarithmic map at the origin. \\
\texttt{retraction(v, x, c)} & First-order approximation to \texttt{expmap}, used by first-order Riemannian optimizers. \\
\texttt{ptransp(v, x, y, c)} & Parallel-transport tangent vector $v$ from the tangent space at $x$ to that at $y$. \\
\texttt{ptransp\_0(v, y, c)} & Parallel transport from the origin's tangent space to $y$'s tangent space. \\
\texttt{tangent\_inner(u, v, x, c)} & Riemannian inner product of two tangent vectors at $x$. \\
\texttt{tangent\_norm(v, x, c)} & Riemannian norm of a tangent vector at $x$. \\
\texttt{egrad2rgrad(grad, x, c)} & Convert a Euclidean gradient into the Riemannian gradient at $x$. \\
\texttt{tangent\_proj(v, x, c)} & Project an ambient vector onto the tangent space at $x$. \\
\texttt{is\_in\_manifold(x, c)} & Check whether $x$ satisfies the manifold's constraint within \texttt{atol}. \\
\texttt{is\_in\_tangent\_space(v, x, c)} & Check whether $v$ is a valid tangent vector at $x$ within \texttt{atol}. \\
\bottomrule
\end{tabular}
\caption{The 19 methods of the \texttt{Manifold} protocol. \texttt{dist}/\texttt{dist\_0} additionally take a trailing static \texttt{version\_idx: int} (manifolds with one implementation accept and ignore it); \texttt{is\_in\_manifold}/\texttt{is\_in\_tangent\_space} take a trailing \texttt{atol: float | None = None}, resolved through \texttt{default\_atol(dtype)} when \texttt{None}.}
\label{tab:protocol}
\end{apptable}

Table~\ref{tab:isometry} lists the exact, distance-preserving isometry maps between the three point-set models shipped in \texttt{hyperbolix.manifolds.isometry\_mappings}. Every off-diagonal pair has both directions implemented; none route through \texttt{logmap}/\texttt{expmap} (which would be lossy and slower). The Hyperboloid$\leftrightarrow$PV pair is a direct map (PV coordinates are exactly the space-like part of the hyperboloid point, with the time coordinate reconstructed from the Lorentz constraint); it is not the composition of the other two maps, which would pass through the Poincar\'e conformal factor and lose accuracy near the ball boundary.

\begin{apptable}
\centering
\begin{tabular}{@{}lll@{}}
\toprule
Model pair & Map & Inverse map \\
\midrule
Poincar\'e $\leftrightarrow$ Hyperboloid & \texttt{poincare\_to\_hyperboloid} & \texttt{hyperboloid\_to\_poincare} \\
Poincar\'e $\leftrightarrow$ ProperVelocity & \texttt{poincare\_to\_pv} & \texttt{pv\_to\_poincare} \\
Hyperboloid $\leftrightarrow$ ProperVelocity (direct) & \texttt{hyperboloid\_to\_pv} & \texttt{pv\_to\_hyperboloid} \\
\bottomrule
\end{tabular}
\caption{Exact isometries between the three point-set manifold models (\texttt{hyperbolix.manifolds.isometry\_mappings}). All three pairs are covered in both directions; the Hyperboloid$\leftrightarrow$ProperVelocity pair is a direct map rather than a composition through the Poincar\'e ball.}
\label{tab:isometry}
\end{apptable}

\section{Initialization Defaults}
\label{app:init}

Table~\ref{tab:init} lists the shipped default initialization for each layer family. Where a family does not reproduce its source paper's initialization, the default is scaled by fan-in or fan-out rather than set to a fixed bound.

\begin{apptable}
\centering
\footnotesize
\begin{tabular}{@{}p{5.0cm}p{4.4cm}p{6.2cm}@{}}
\toprule
Layer family & Default init & Rationale \\
\midrule
Poincar\'e FC (\texttt{HypLinearPoincare}, \texttt{HypLinearPoincarePP}) & $\mathrm{std} = 1/\sqrt{\text{fan\_in}}$ & Keeps first-layer output away from the ball boundary. \\
Poincar\'e regression / MLR (\texttt{HypRegressionPoincare}, \texttt{HypRegressionPoincarePP}) & $\mathrm{std} = (2\cdot\text{in}\cdot\text{out})^{-1/2}$ & Matches \citep{vanspengler2023resnet}; unscaled $\mathcal{N}(0,1)$ row norms $\approx\sqrt{\text{in\_dim}}$ overwhelm the MLR output scaling. \\
Poincar\'e conv (\texttt{HypConv2DPoincare}, \texttt{id\_init=True}) & $0.5\cdot\mathbf{I}$ (identity) & The $0.5$ factor compensates the factor of $2$ inside the HNN++ distance formula. \\
\texttt{HTCLinear} (+ wrappers, e.g.\ attention Q/K/V, \texttt{HypformerPositionalEncoding}) & fan-in uniform $U(\pm\sqrt{3/\text{in\_features}})$ & Per-layer input-Jacobian gain $\approx 1$ (no nonlinearity in the \texttt{htc} tail); a fixed bound (e.g.\ the legacy $0.02$) is width-dependent and contracts at large fan-in. \\
FHNN / FHCNN linear \& conv (\texttt{HypLinearHyperboloidFHNN}, \texttt{HypLinearHyperboloidFHCNN}, \texttt{HypConv2DHyperboloidFHNN}) & $U(-0.02, 0.02)$ & Matches \citep{chen2022fhnn,bdeir2024fhcnn} reference init (FHNN additionally zeroes the time column, giving tangent vectors at the origin). \\
PLFC / ILNN linear (\texttt{HypLinearHyperboloidPLFC}) & $\mathrm{std} = 0.02$ & Matches the \citep{shi2026ilnn} PLFC reference value. \\
ILNN conv (\texttt{HypConv2DHyperboloidILNN}) & $\mathrm{std} = \sqrt{1/\text{out\_spatial}}$ (fan-out, default) & Norm-preserving under the corrected LogCat digamma rescale; \texttt{kernel\_init\_std=0.02} restores the paper reference (tuned for the pre-fix LogCat and now strongly contractive). \\
PV hidden layers (\texttt{HypLinearPV}, \texttt{HypConv2DPV}) & He, $\sqrt{2/\text{fan\_in}}$ & Near the origin the PV output reduces to a Euclidean linear map, so standard He/ReLU variance-preservation analysis applies. \\
PV regression head (\texttt{HypRegressionPV}) & $\mathrm{std} = 10^{-2}$ & Matches the paper's \texttt{PVManifoldMLR.reset\_parameters}, tuned for $O(1)$-variance input features immediately before a softmax/MSE loss. \\
FGG linear \& conv, default (\texttt{FGGLinear}, \texttt{FGGConv2D}, \texttt{reset\_params="fan\_out"}) & $\mathrm{std} = \sqrt{1/\text{out\_spatial}}$ & Norm-preserving; a deliberate RL-tuned deviation from the \citep{vanDerKlis2026fgglnn} classification reference (\texttt{"eye"} init + \texttt{init\_bias=0.5}), which assumes a following BatchNorm. \\
Hyperboloid HCat conv, default (\texttt{HypConv2DHyperboloid}, \texttt{reset\_params="default"}) & $U(-0.02, 0.02)$ & Unchanged FHCNN-style init; alternative \texttt{"fan\_in"} uses $\mathrm{std}=\sqrt{1/\text{hcat\_out\_ambient\_dim}}$. \\
Busemann FC (\texttt{HypLinearHyperboloidBusemann}, \texttt{HypLinearPoincareBusemann}) & weight-norm split, direction $\mathrm{std}=(2\cdot\text{in}\cdot\text{out})^{-1/2}$, $\log\_\mathrm{scale}=\log\lVert\text{kernel row}\rVert$ & Reproduces the reference \texttt{weight\_g}/\texttt{weight\_v} split; matches the HNN++ FC scale convention. \\
Busemann MLR (\texttt{HypRegressionHyperboloidBusemann}, \texttt{HypRegressionPoincareBusemann}) & direction $\mathrm{std}=\text{in\_spatial}^{-1/2}$, $\log\_\mathrm{scale}=\log\lVert\text{kernel row}\rVert$ & Same weight-norm split as the FC layers, matching the paper's MLR scale. \\
\bottomrule
\end{tabular}
\caption{Shipped initialization defaults by layer family. Also the reference values for custom reimplementations.}
\label{tab:init}
\end{apptable}

Across every family above, the shipped default is chosen so the per-layer input-Jacobian gain sits near $1$. This single design constraint has an asymmetric failure mode on either side. A generic Euclidean init (He/Xavier, or a fixed bound calibrated for a different fan-in) that overshoots gain $1$ pushes the first-layer output toward the Poincar\'e boundary or far up the hyperboloid, producing \texttt{NaN} losses within a few steps — loud and easy to diagnose. An init that undershoots gain $1$ fails silently: for a stack of linear-in-the-matmul layers the gain compounds as $g^{\text{depth}}$, and once pairwise output distances fall below the float32 resolution of the distance computation itself (near the origin, \texttt{acosh}$(1+c\,d^2/2)$ quantizes to exactly zero once $c\,d^2/2 < \varepsilon_{f32}\approx1.19\times10^{-7}$), the stack becomes a constant map: gradients are $\approx 0$ from step zero, with no \texttt{NaN} or warning to flag it. Both failure modes have been observed in this codebase's own history (documented in the \texttt{HTCLinear} and \texttt{HypConv2DHyperboloidILNN} docstrings) and motivate replacing library-wide fixed-bound inits with fan-in/fan-out-scaled ones.

\section{Numerical Conventions}
\label{app:numerics}

\paragraph{\texttt{MIN\_NORM} and its finite-VJP rationale.} \texttt{hyperbolix.utils.math\_utils.MIN\_NORM = 1e-15} is the library-wide gradient-safety floor for norms and denominators, stabilizing gradients in two ways. First, \texttt{safe\_norm} (a max-scaled reduction floored at \texttt{MIN\_NORM} via \texttt{floor\_at}) has a finite gradient at $x=0$. In contrast, a plain \texttt{jnp.linalg.norm} evaluates the derivative as $x/\lVert x\rVert = 0/0 = \mathrm{NaN}$. Second, \texttt{maximum(denom, MIN\_NORM)} keeps a division finite when a denominator collapses. $10^{-15}$ is small enough to be invisible next to any float32 or float64 quantity the library works with (float64 machine epsilon is $2.2\times10^{-16}$).

\paragraph{\texttt{tanh} input/output clamping.} \texttt{hyperbolix.utils.math\_utils.tanh} clips its \emph{input} to $\pm0.5\log(2/\varepsilon)$ ($\approx\pm 8.3$ for float32, $\approx\pm 18.4$ for float64) and additionally clips its \emph{output} to $\pm(1-10\varepsilon)$. Both clips are needed: the input clip is the analytic point where $1-\tanh(x)$ reaches $\varepsilon$, but XLA's float32 \texttt{tanh} saturates to \emph{exactly} $1.0$ slightly before that bound (around $x=8$), which the input clip alone does not prevent. An output of exactly $1.0$ would make a downstream \texttt{atanh} call hit its pole and inject a \texttt{NaN} cotangent into the VJP that no later \texttt{jnp.where} can remove. The output clip matches \texttt{atanh}'s own domain guard, so \texttt{atanh(tanh(x))} can never reach the singularity.

\paragraph{Tolerance conventions.} Table~\ref{tab:tol} gives the two absolute-tolerance conventions used throughout the test suite and library.

\begin{apptable}
\centering
\begin{tabular}{@{}lll@{}}
\toprule
Convention & float32 & float64 \\
\midrule
\texttt{default\_atol(dtype)} $= \sqrt{\varepsilon_{\mathrm{dtype}}}$ (manifold membership) & $3.45\times10^{-4}$ & $1.49\times10^{-8}$ \\
Suite-wide dtype fixture \texttt{atol}/\texttt{rtol} (\texttt{tests/conftest.py}) & $4\times10^{-3}$ & $1\times10^{-7}$ \\
\bottomrule
\end{tabular}
\caption{Default absolute tolerances. \texttt{default\_atol} is the resolution used by every \texttt{is\_in\_manifold}/\texttt{is\_in\_tangent\_space} call left at \texttt{atol=None}; an explicit \texttt{atol} always wins and is never floored or clamped.}
\label{tab:tol}
\end{apptable}

\texttt{default\_atol} is $\sqrt{\varepsilon}$ because the constraint residuals it checks---$\langle x,x\rangle_{\mathcal{L}} + 1/c$ on the hyperboloid, $c\lVert x\rVert^2 - 1$ on the ball---are differences of same-magnitude sums that lose roughly half the mantissa. Both residuals are dimensionless, so a single tolerance works at every curvature. The tolerance is absolute and calibrated for points near the origin: a genuinely on-manifold point far from the origin can fail the default check (float64 holds to geodesic distance ${\sim}10$, float32 to ${\sim}7$), which is the regime where the library recommends float64.

\paragraph{Dtype guidance.} Table~\ref{tab:dtype} summarizes when float32 is sufficient.

\begin{apptable}
\centering
\begin{tabular}{@{}lll@{}}
\toprule
Geodesic distance from origin & float32 accuracy & Recommendation \\
\midrule
$d < 3$ & excellent ($<0.01\%$ error) & float32 \\
$3 \le d < 5$ & good ($<0.1\%$ error) & float32 \\
$5 \le d < 10$ & moderate ($<3\%$ error) & float64 for critical ops \\
$d \ge 10$ & poor ($>3\%$ error) & float64 required \\
\bottomrule
\end{tabular}
\caption{Poincar\'e ball precision requirements by geodesic distance from the origin. The hyperboloid works for float32 much further out due to its cancellation-free formulas.}
\label{tab:dtype}
\end{apptable}

Storage and compute precision are separate, orthogonal knobs. \emph{Compute} precision is the dtype in which manifold operations (\texttt{dist}, \texttt{expmap}, \texttt{logmap}, \ldots) run, set by the manifold instance's \texttt{dtype} attribute (e.g.\ \texttt{Poincare(dtype=jnp.float64)}); manifold methods cast their array arguments to this dtype on entry. \emph{Storage} precision is the dtype of a layer's trainable parameters and persistent state (batch-norm running statistics, VQ codebooks), set independently by the \texttt{param\_dtype} constructor argument on every NN layer (default \texttt{jnp.float32}). A float32-stored parameter that enters a float64 manifold computation is promoted for that computation only; the stored parameter itself is untouched. The recommended high-precision recipe is therefore float64 compute with float32 storage — full geometric precision where it matters, at half the parameter/optimizer-state memory.

\section{Test-Suite Structure}
\label{app:tests}

At v1.3.0 the suite comprises 55 \texttt{test\_*.py} files (two \texttt{conftest.py} fixture files and two \texttt{\_\_init\_\_.py} files bring the directory to 59 \texttt{.py} files) totaling 1{,}038 top-level \texttt{def test\_*} functions. Under \texttt{pytest} parametrization (dtype $\times$ curvature $\times$ layer-spec axes) this expands to 4{,}446 collected test items.

\subsection{The 18-property layer contract}

\texttt{tests/nn\_layers/test\_layer\_contract.py} replaces the former per-layer six-slot boilerplate (copy-pasted once per layer class across a dozen files) with a single \texttt{LayerSpec} table and five parametrized contract tests, run over every layer in the table. Enumerated as 18 individually checkable properties:

\begin{enumerate}
\item Forward output shape matches the declared \texttt{out\_shape}.
\item Forward output dtype matches the input dtype (float32).
\item Forward output is entirely finite.
\item Forward output lies on the manifold (\texttt{is\_in\_manifold}), for every manifold-valued layer.
\item \texttt{nnx.jit}-compiled forward output shape matches the eager result.
\item \texttt{nnx.jit}-compiled forward output values match the eager result (\texttt{allclose}, $\mathrm{atol}=\mathrm{rtol}=10^{-5}$).
\item Output is finite at a non-unit curvature $c=0.5$.
\item Output stays on the manifold at $c=0.5$.
\item Output is finite at a non-unit curvature $c=2.0$.
\item Output stays on the manifold at $c=2.0$.
\item Loss (sum of squared outputs) is finite.
\item Every declared gradient path (\texttt{kernel}, \texttt{bias}, and any layer-specific extra parameter such as \texttt{scale}, \texttt{log\_scale}, \texttt{gyro\_bias}) is finite, at both float32 and float64.
\item The primary parameter's gradient has at least one nonzero entry (non-trivial signal).
\item \texttt{nnx.jit}-compiled gradient equals the eager gradient for every declared path (tolerance scaled by the largest gradient entry).
\item The \texttt{input\_space="tangent"} forward output has the correct shape.
\item The \texttt{input\_space="tangent"} branch agrees exactly with an explicit \texttt{expmap\_0} followed by the \texttt{input\_space="manifold"} branch ($\mathrm{atol}=10^{-9}$, float64).
\item Convolutional output spatial size follows the declared padding/stride arithmetic, across kernel size $\times$ padding $\times$ stride combinations.
\item Convolutional output is entirely finite under every kernel/padding/stride combination.
\end{enumerate}

\subsection{Oracle taxonomy}

Every numeric assertion in the suite is checked against one of three independent oracle sources — never against another hyperbolic-geometry library:
\begin{itemize}
\item \textbf{Hand-derived closed forms}, worked out independently of the implementation and, where cancellation is a concern, evaluated at extended precision (e.g.\ \texttt{numpy.longdouble}, 80-bit on x86) so the oracle's own rounding error sits several decades below the assertion tolerance.
\item \textbf{NumPy/SciPy transcriptions} of the source papers' formulas, built from primitive NumPy operations rather than calling into any hyperbolic-geometry package (e.g.\ \texttt{scipy.stats.multivariate\_normal} for the flat-curvature limit of the wrapped normal, or a from-scratch NumPy geodesic frame for the log-probability integration test).
\item \textbf{Finite differences / independent autodiff}, e.g.\ cross-checking a closed-form log-Jacobian against \texttt{jax.jacfwd} of the sampling map itself, so a bug shared between the closed form and its \emph{use} would still be caught by the independent differentiation path.
\end{itemize}

\subsection{Wrapped-normal log-probability: four-direction pinning}

\texttt{tests/test\_wrapped\_normal.py} pins the wrapped-normal \texttt{log\_prob} value (not just its shape/finiteness) from four independent directions, for both the Hyperboloid and Poincar\'e implementations:
\begin{enumerate}
\item \textbf{Quadrature integrates to one.} $\int \exp(\texttt{log\_prob})\,dV = 1$ against the Riemannian volume element of $\mathbb{H}^2$ in geodesic polar coordinates (Gauss-Legendre in radius, trapezoid in angle), at three curvatures and a non-origin mean.
\item \textbf{Change-of-variables vs.\ \texttt{jacfwd}.} \texttt{log\_prob} is checked against the pushforward density of the sampling map itself: the tangent-space Gaussian density divided by $\lvert\det \partial T/\partial v\rvert$, with the Jacobian obtained from \texttt{jax.jacfwd} of the sampling map — never from the library's own closed-form log-determinant.
\item \textbf{Flat limit vs.\ SciPy multivariate normal.} At $c=10^{-4}$, both models must reduce to a Euclidean Gaussian in Riemannian coordinates, matching \texttt{scipy.stats.multivariate\_normal.logpdf} to $10^{-3}$ (the residual is the expected $O(c\,r^2)$ curvature correction).
\item \textbf{Isometry cross-check.} The Hyperboloid and Poincar\'e log-densities must agree \emph{pointwise, with no correction factor} at corresponding points related by \texttt{hyperboloid\_to\_poincare} — since that map is an isometry, it carries the Riemannian volume measure over unchanged.
\end{enumerate}

\clearpage
\section{Measured Accuracy, Stability, and Speed}
\label{app:benchmarks}

This appendix reports the measurements behind Table~\ref{tab:oracle}, Figures~\ref{fig:crosslib_stability}, \ref{fig:roundtrip_stability} and~\ref{fig:cuda_speed}, and the speed and cancellation tables that follow. Every number comes from the released wheel hyperbolix 1.3.0, installed from PyPI into a fresh environment per script rather than imported from a source checkout, so the tables describe what a user installs. The baselines are geoopt 0.5.1~\citep{kochurov2020geoopt} and HypLL 0.1.1~\citep{vanspengler2023hypll} on PyTorch 2.13.0. The six measurement scripts write JSON, and five rendering scripts turn that JSON into every table and figure without touching any library. 

\subsection{Common protocol}
\label{app:measure_protocol}

\paragraph{Oracles.} Every accuracy number is a deviation from a NumPy float64 transcription of a published closed form, evaluated on points that are themselves sampled in NumPy. No oracle calls hyperbolix, geoopt, or HypLL; hyperboloid points are lifted with the NumPy $\exp_0$ oracle rather than with any library map. A deviation is therefore always library against reference, never library against itself.

\paragraph{Timing.} All speed measurements use $D = 32$ spatial dimensions, curvature $c = 1$, and float32. Operands are generated once in NumPy and handed to every library, so the columns of a row time the same numbers, not merely the same shapes. Every timed callable synchronizes internally (\texttt{jax.block\_until\_ready} for JAX, \texttt{torch.cuda.synchronize} for PyTorch on GPU); the timer holds only the clock. A cell is the median over 20 timed repeats after 5 warm-up calls. HypLL's \texttt{ManifoldTensor} wrappers are constructed outside the timed region. The hyperbolix eager column runs \texttt{jax.vmap} over the single-point API without \texttt{jax.jit}, which re-traces on every call; it measures host-side dispatch, and the jit column is the kernel comparison.

\paragraph{Launches and brackets.} We call one process invocation of the speed script a \emph{launch}. Every speed table aggregates five sequential launches: a cell is the median of the five per-launch medians, and the bracket beneath it is the minimum and maximum of those five medians. All five launches use the same seed and therefore identical operands, so a bracket bounds launch-to-launch machine and compilation state, not sampling variation. Where we report that one library is faster ``in all five launches'', the ordering held across all launches, and the two brackets do not overlap.

\paragraph{Absent cells.} An n/a cell is structurally absent rather than unmeasured, with one exception. HypLL ships no hyperboloid model, so its hyperboloid cells do not exist; geoopt ships no neural-network layers, so it has no layer rows. The exception is the HypLL compiled cell of the Poincar\'e linear row: HypLL's \texttt{HLinear} holds a \texttt{ManifoldParameter} whose \texttt{\_\_torch\_function\_\_} raises when Dynamo probes it, so \texttt{torch.compile} fails on that layer while succeeding on all five HypLL primitives. Two rows are not projection-matched: HypLL's \texttt{mobius\_add} does not project its result back into the ball, whereas geoopt's and hyperbolix's do.

\subsection{Deviation from independent oracles}
\label{app:oracle}

Table~\ref{tab:oracle} measures hyperbolix alone against the oracles. For every manifold, operation class, and dtype, the script samples 2\,048 points per cell over dimensions $\{2, 10\}$ and curvatures $\{0.3, 1.0, 2.5\}$; the $\kappa$-stereographic model adds the spherical branch $c = -1$, on which tangent norms are capped at $0.9 \cdot \pi/2$, below the pole of $\tan$. Each table row aggregates the maximum and the median absolute deviation over its whole curvature and dimension grid. The layer-forward row (MLR) exists for the hyperboloid and proper-velocity manifolds, which are the two on which the library ships a closed-form multinomial regression head with a published oracle.

Every float32 cell lies below $4.6 \times 10^{-6}$ and every float64 cell at or below $3.5 \times 10^{-14}$, so all cells satisfy the library's membership tolerance $\sqrt{\varepsilon}$ ($3.45 \times 10^{-4}$ in float32, $1.49 \times 10^{-8}$ in float64) with margin. The deviations are absolute, which matters for reading the largest float32 cell: stereographic distance, at $4.6 \times 10^{-6}$, is only 1.15$\times$ the next-largest float32 cell (Poincar\'e distance, $4.0 \times 10^{-6}$). The oracle itself is a fresh derivation; it satisfies the tangency invariant $\langle y, \mathrm{PT}_{x \to y} v \rangle_{\mathcal{L}} = 0$ to $2.8 \times 10^{-14}$ over the whole grid.

\begin{apptable}
    \centering
    \small
    \caption{Maximum and median \emph{absolute} deviation between hyperbolix 1.3.0 operations and independent NumPy float64 oracles, over 2\,048 sampled points per (curvature, dimension) cell, curvatures $\{0.3, 1.0, 2.5\}$ (plus $c = -1$ for the stereographic model) and dimensions $\{2, 10\}$. Every cell lies below the membership tolerance $\sqrt{\varepsilon}$ of its dtype. The largest float32 cell (stereographic distance, $4.6\times10^{-6}$) is only $1.15\times$ the next-largest cell (Poincar\'e distance). Layer-forward (MLR) rows exist for the hyperboloid and proper-velocity manifolds only. Script: \texttt{scripts/oracle\_deviation.py}, CPU backend.}
    \label{tab:oracle}
    \begin{tabular}{llcccc}
      \toprule
      & & \multicolumn{2}{c}{float32} & \multicolumn{2}{c}{float64} \\
      \cmidrule(lr){3-4} \cmidrule(lr){5-6}
      Manifold & Operation class & max & median & max & median \\
      \midrule
    Poincar\'e & dist & $4.0\times10^{-6}$ & $1.6\times10^{-7}$ & $7.1\times10^{-15}$ & $3.3\times10^{-16}$ \\
    Poincar\'e & expmap/logmap & $2.7\times10^{-7}$ & $5.8\times10^{-9}$ & $4.4\times10^{-16}$ & $0$ \\
    Poincar\'e & ptransp & $4.3\times10^{-7}$ & $4.4\times10^{-9}$ & $6.7\times10^{-16}$ & $0$ \\
    \midrule
    Hyperboloid & dist & $1.1\times10^{-6}$ & $7.6\times10^{-8}$ & $3.5\times10^{-14}$ & $2.2\times10^{-16}$ \\
    Hyperboloid & expmap/logmap & $1.4\times10^{-6}$ & $8.6\times10^{-9}$ & $4.7\times10^{-15}$ & $0$ \\
    Hyperboloid & ptransp & $1.4\times10^{-6}$ & $7.2\times10^{-9}$ & $3.6\times10^{-15}$ & $0$ \\
    Hyperboloid & MLR forward & $2.0\times10^{-6}$ & $4.1\times10^{-8}$ & $3.6\times10^{-15}$ & $1.0\times10^{-17}$ \\
    \midrule
    ProperVelocity & dist & $1.3\times10^{-6}$ & $9.6\times10^{-8}$ & $8.2\times10^{-15}$ & $2.2\times10^{-16}$ \\
    ProperVelocity & expmap/logmap & $1.1\times10^{-6}$ & $1.3\times10^{-8}$ & $2.7\times10^{-15}$ & $0$ \\
    ProperVelocity & ptransp & $1.4\times10^{-6}$ & $1.0\times10^{-8}$ & $1.8\times10^{-15}$ & $0$ \\
    ProperVelocity & MLR forward & $1.9\times10^{-6}$ & $6.3\times10^{-8}$ & $2.7\times10^{-15}$ & $0$ \\
    \midrule
    Stereographic & dist & $4.6\times10^{-6}$ & $1.2\times10^{-7}$ & $1.1\times10^{-14}$ & $2.2\times10^{-16}$ \\
    Stereographic & expmap/logmap & $3.8\times10^{-6}$ & $6.0\times10^{-9}$ & $6.2\times10^{-15}$ & $0$ \\
    Stereographic & ptransp & $4.6\times10^{-7}$ & $4.6\times10^{-9}$ & $4.4\times10^{-16}$ & $0$ \\
      \bottomrule
    \end{tabular}
\end{apptable}

\subsection{Cross-library accuracy on the Poincar\'e ball}
\label{app:crosslib_accuracy}

Table~\ref{tab:crosslib_accuracy} applies the same oracles to all three libraries. It is restricted to the Poincar\'e ball because that is the only model all three implement. For each of five operations, three curvatures $\{0.3, 1.0, 2.5\}$, two dimensions $\{2, 10\}$, and both dtypes, the script samples 2\,048 points, evaluates each library and the oracle on the same points, and records the maximum absolute deviation; the table reports the worse of the two dimensions.

In float32 the three libraries are interchangeable: across the 90 (operation, curvature, dimension, library) cells before the dimension collapse, deviations lie between $6.3 \times 10^{-8}$ and $4.8 \times 10^{-6}$, and within any one (operation, curvature, dimension) cell the three libraries differ by at most 2.2$\times$ (median 1.24$\times$). The float64 half separates them, and the separation is about curvature storage rather than arithmetic. HypLL stores its curvature as float32 (\texttt{Curvature.\_\_init\_\_} casts to \texttt{torch.float32}, and \texttt{.double()} cannot recover the lost digits), so at $c = 0.3$ every HypLL float64 table cell floors between $1.8 \times 10^{-8}$ and $8.8 \times 10^{-8}$ (over the underlying grid, before the dimension collapse, $1.6 \times 10^{-8}$ to $8.8 \times 10^{-8}$), while $c = 1$ and $c = 2.5$, which are exactly representable in float32, reach machine precision. geoopt keeps the inverse softplus of $c$ as a parameter and passes a Python float through \texttt{torch.as\_tensor}, which also lands in float32; the natural call \texttt{PoincareBall(c=0.3)} floors at ${\sim}10^{-7}$, worse than HypLL (measured in a side probe of the accuracy audit; the released grid contains only the typed-tensor call). The geoopt column reaches machine precision only because the script passes an explicitly typed float64 tensor, an escape hatch HypLL lacks. hyperbolix takes the curvature as a call-time scalar in the working dtype and has no such floor.

\begin{apptable}
    \centering
    \footnotesize
    \setlength{\tabcolsep}{4pt}
    \caption{Maximum absolute deviation from the NumPy float64 oracle on the Poincar\'e ball, for geoopt 0.5.1, HypLL 0.1.1, and hyperbolix 1.3.0, over 2\,048 sampled points per cell and the worse of dimensions $\{2, 10\}$. In float32 the three libraries agree to within 2.2$\times$ in every cell. In float64, HypLL's cells at $c = 0.3$ floor at ${\sim}10^{-8}$ because HypLL stores the curvature in float32; geoopt reaches machine precision only when given an explicitly typed float64 tensor rather than a Python float. Script: \texttt{scripts/crosslib\_accuracy.py}, CPU backend.}
    \label{tab:crosslib_accuracy}
    \begin{tabular}{llcccccc}
      \toprule
      & & \multicolumn{3}{c}{float32} & \multicolumn{3}{c}{float64} \\
      \cmidrule(lr){3-5} \cmidrule(lr){6-8}
      Operation & $c$ & geoopt & HypLL & hyperbolix & geoopt & HypLL & hyperbolix \\
      \midrule
    $d(x, y)$ & 0.3 & $4.8\times10^{-6}$ & $3.8\times10^{-6}$ & $4.0\times10^{-6}$ & $3.6\times10^{-14}$ & $8.8\times10^{-8}$ & $8.8\times10^{-15}$ \\
    $d(x, y)$ & 1.0 & $1.7\times10^{-6}$ & $1.6\times10^{-6}$ & $1.7\times10^{-6}$ & $6.7\times10^{-15}$ & $3.1\times10^{-15}$ & $3.1\times10^{-15}$ \\
    $d(x, y)$ & 2.5 & $1.2\times10^{-6}$ & $1.2\times10^{-6}$ & $1.5\times10^{-6}$ & $3.3\times10^{-15}$ & $2.4\times10^{-15}$ & $2.9\times10^{-15}$ \\
    \midrule
    $\exp_0(v)$ & 0.3 & $3.5\times10^{-7}$ & $2.5\times10^{-7}$ & $2.6\times10^{-7}$ & $3.1\times10^{-15}$ & $3.0\times10^{-8}$ & $6.7\times10^{-16}$ \\
    $\exp_0(v)$ & 1.0 & $1.0\times10^{-7}$ & $1.1\times10^{-7}$ & $1.5\times10^{-7}$ & $5.6\times10^{-16}$ & $3.3\times10^{-16}$ & $2.2\times10^{-16}$ \\
    $\exp_0(v)$ & 2.5 & $9.8\times10^{-8}$ & $8.1\times10^{-8}$ & $9.8\times10^{-8}$ & $2.2\times10^{-16}$ & $2.2\times10^{-16}$ & $2.2\times10^{-16}$ \\
    \midrule
    $\log_0(x)$ & 0.3 & $5.1\times10^{-7}$ & $2.6\times10^{-7}$ & $2.7\times10^{-7}$ & $2.0\times10^{-15}$ & $1.8\times10^{-8}$ & $4.4\times10^{-16}$ \\
    $\log_0(x)$ & 1.0 & $1.3\times10^{-7}$ & $1.3\times10^{-7}$ & $1.4\times10^{-7}$ & $5.6\times10^{-16}$ & $2.2\times10^{-16}$ & $2.2\times10^{-16}$ \\
    $\log_0(x)$ & 2.5 & $9.8\times10^{-8}$ & $7.7\times10^{-8}$ & $7.7\times10^{-8}$ & $2.2\times10^{-16}$ & $1.7\times10^{-16}$ & $1.1\times10^{-16}$ \\
    \midrule
    $x \oplus_c y$ & 0.3 & $3.9\times10^{-7}$ & $4.9\times10^{-7}$ & $4.8\times10^{-7}$ & $9.0\times10^{-16}$ & $3.1\times10^{-8}$ & $8.9\times10^{-16}$ \\
    $x \oplus_c y$ & 1.0 & $2.2\times10^{-7}$ & $2.2\times10^{-7}$ & $2.7\times10^{-7}$ & $2.2\times10^{-16}$ & $2.2\times10^{-16}$ & $5.6\times10^{-16}$ \\
    $x \oplus_c y$ & 2.5 & $1.3\times10^{-7}$ & $1.3\times10^{-7}$ & $1.6\times10^{-7}$ & $1.7\times10^{-16}$ & $1.7\times10^{-16}$ & $4.4\times10^{-16}$ \\
    \midrule
    $PT_{0 \to x}(v)$ & 0.3 & $6.3\times10^{-7}$ & $4.3\times10^{-7}$ & $4.3\times10^{-7}$ & $8.9\times10^{-16}$ & $7.0\times10^{-8}$ & $4.4\times10^{-16}$ \\
    $PT_{0 \to x}(v)$ & 1.0 & $2.2\times10^{-7}$ & $2.2\times10^{-7}$ & $2.2\times10^{-7}$ & $3.3\times10^{-16}$ & $3.3\times10^{-16}$ & $2.2\times10^{-16}$ \\
    $PT_{0 \to x}(v)$ & 2.5 & $1.5\times10^{-7}$ & $1.5\times10^{-7}$ & $1.6\times10^{-7}$ & $2.2\times10^{-16}$ & $2.2\times10^{-16}$ & $2.2\times10^{-16}$ \\
      \bottomrule
    \end{tabular}
\end{apptable}

\subsection{Round-trip stability}
\label{app:stability_measure}

Figures~\ref{fig:crosslib_stability} and~\ref{fig:roundtrip_stability} share one setup: dimension 8, $c = 1$, 512 tangent vectors per radius, and 40 norms spaced logarithmically from $10^{-3}$ to $20$. The $x$ axis is the tangent-vector norm (which equals geodesic distance on the hyperboloid and half the geodesic distance on the Poincar\'e ball). Both ran on the CPU backend, with relative errors computed in float64.

We define the \emph{breakdown radius} as the first grid point where the float32 median relative error exceeds $\sqrt{\varepsilon_{32}}$. On the Poincar\'e ball it is $r = 3.38$ for geoopt and HypLL and $r = 7.24$ for hyperbolix. This difference is due to a design choice about the projection margin: hyperbolix projects onto a ball of radius $1/\sqrt{c} - \varepsilon^{0.75}$, and $\operatorname{artanh}(1 - \varepsilon_{32}^{0.75}) = 6.33$ is where the round trip must fail. On the hyperboloid, geoopt already exceeds the threshold at $r = 10^{-3}$ with a median of $2.3 \times 10^{-2}$ against $1.1 \times 10^{-7}$ for hyperbolix. hyperbolix stays below it through $r = 20$.

The hyperbolix-only figure adds the proper-velocity and $\kappa$-stereographic models, both dtypes, and the distance variants selected through \texttt{version\_idx}. Its float32 Poincar\'e curve has a knee at tangent norm $r = 7.24$ (geodesic distance 14.5), consistent with the margin above; in float64 the same round trip is still $2.8 \times 10^{-8}$ at $r = 12.0$. The proper velocity to hyperboloid isometry round trip is exactly zero in both dtypes by construction: the map to the hyperboloid concatenates a recomputed time coordinate, and the map back slices it off. Near the origin, the hyperboloid \texttt{dist\_0} default and the $\log_0(\exp_0(v))$ round trip track $r$ down to $r = 10^{-3}$ with a float32 relative error between 0 and $2.4 \times 10^{-7}$; the smoothened distance variant (\texttt{VERSION\_SMOOTHENED}), which floors the spatial radius in quadrature so that the distance is never exactly zero and its gradient stays finite through coincidence, has a relative error of $2.7 \times 10^{-6}$ at $r = 10^{-3}$ that decays with $r$ ($1.6 \times 10^{-6}$, $1.1 \times 10^{-6}$, $4.3 \times 10^{-7}$ at the next three grid points), a peak rather than a floor.

\subsection{Speed on an NVIDIA H100}
\label{app:speed_cuda}

Tables~\ref{tab:speed} and~\ref{tab:speed_cuda_eager} give the compiled and the eager columns of the CUDA benchmark; Figure~\ref{fig:cuda_speed} plots the compiled columns at batch $10^7$. The five launches ran back to back as one SLURM job array (\texttt{speed\_cuda\_130}) on a single NVIDIA H100 (driver 575.57.08) of a shared node that was not reserved exclusively, with Python 3.13.7, JAX 0.11.1, and PyTorch 2.13.0 (CUDA 12.9 wheel), over batch sizes $10^2$ to $10^7$. The tables print $10^3$, $10^5$, and $10^7$. An H100 is dispatch-bound on these operations up to about $10^5$ points: the HypLL compiled column sits at a flat 0.05--0.08\,ms from $10^2$ to $10^5$, the geoopt compiled column between 0.06 and 0.43\,ms, and the hyperbolix jit column at a median of about 0.07\,ms (per-operation 0.06--0.09\,ms at $10^2$), nearly flat across batch size, which is why the sweep extends to $10^7$ and the main-text figure plots the largest batch.

At batch $10^7$, compiled against compiled, the picture splits by manifold. On the Poincar\'e ball, HypLL's kernels are fastest on
distance, $\exp_0$, and $\log_0$ in all five launches (1.6--2.3$\times$ ahead of hyperbolix, 3.1--12.6$\times$ ahead of geoopt) \footnote{For \texttt{dist} and \texttt{logmap0}, a data-dependent branch in geoopt's \texttt{artan\_k} prevents \texttt{torch.compile} from tracing the graph, so the compiled column silently falls back to eager mode. The 12.6$\times$ and 3.1$\times$ ratios for these two operations therefore compare HypLL's compiled kernel against
geoopt's host dispatch.}. The two PyTorch libraries are level on M\"obius addition and parallel transport, both roughly 2.5$\times$ faster than hyperbolix. On the hyperboloid, hyperbolix is 1.9--2.9$\times$ faster than geoopt on $\exp_0$, $\log_0$, and parallel transport. The distance has no stable ordering as geoopt's results are bimodal across launches. The gap in the Poincar\'e ball operations is a compiler limitation, not an algorithmic one. Each of these operations reduces a row to a scalar (e.g.\ a norm), then rescales the same row by a function of that scalar. A single GPU kernel could perform both steps in a single pass over the data. XLA instead splits the reduction and the rescale into separate kernels, forcing the full $(B, D)$ input to be written out and read back between them. This explains why hyperbolix loses the most in M\"obius addition and parallel transport: they have the most reductions per call, and are $2.55\times$ and $2.45\times$ slower than HypLL (Table~\ref{tab:speed}). HLO-level profiling on an A100 corroborates the fusion explanation: the measured HBM-byte ratios for these two
operations are $2.57\times$ and $2.45\times$, closely tracking the wall-clock gap. In contrast, hyperbolix is competitive for $\log_0$ because it has only one reduction and the smallest overhead (1.56$\times$). The two-layer rows are hyperbolix-only: geoopt ships no layers, and HypLL's layer cannot be compiled.

The eager columns (Table~\ref{tab:speed_cuda_eager}) compare per-call overhead, not kernels. The hyperbolix eager cell is nearly flat from $10^3$ to $10^5$ (M\"obius addition: 19.8 and 20.4\,ms) and rises only at $10^7$ (33.6\,ms), because re-tracing \texttt{vmap} on every call dominates until the kernel itself takes longer than the trace. The PyTorch eager columns rise with batch as expected. A speed-up quoted against the hyperbolix eager column is therefore a Python-overhead ratio; the compiled table is the kernel comparison.

\begin{apptable}
    \centering
    \small
    \caption{Compiled wall-clock time per call (ms) on one NVIDIA H100 for matched primitives and layers, float32, $D = 32$, $c = 1$: geoopt and HypLL under \texttt{torch.compile}, hyperbolix under \texttt{jax.jit(jax.vmap($\cdot$))}. Cell = median of five launch medians, each over 20 repeats; bracket = min--max across the five launches, which share one seed and so bound launch-to-launch state, not sampling. The batches $10^2$, $10^4$, and $10^6$ are in the JSON artifact. n/a cells are structural (HypLL has no hyperboloid, geoopt no layers) except the HypLL Poincar\'e linear cell, where \texttt{torch.compile} fails on HypLL's \texttt{ManifoldParameter}. The hyperboloid \texttt{dist} geoopt cell is bimodal across launches (4.37--4.40\,ms in four, 10.3\,ms in one). Script: \texttt{scripts/speed\_benchmark.py --device cuda}.}
    \label{tab:speed}
    \begin{tabular}{llccc}
      \toprule
      Operation & Batch & geoopt (compile) & HypLL (compile) & hyperbolix (jit) \\
      \midrule
    Poincar\'e \texttt{dist} & $10^{3}$ & 0.282 {\scriptsize[0.276, 0.393]} & 0.062 {\scriptsize[0.059, 0.065]} & 0.079 {\scriptsize[0.063, 0.098]} \\
     & $10^{5}$ & 0.387 {\scriptsize[0.337, 0.561]} & 0.068 {\scriptsize[0.062, 0.106]} & 0.081 {\scriptsize[0.066, 0.101]} \\
     & $10^{7}$ & 15.4 {\scriptsize[15.4, 15.5]} & 1.22 {\scriptsize[1.22, 1.24]} & 2.76 {\scriptsize[2.71, 2.77]} \\
    Poincar\'e $\exp_0$ & $10^{3}$ & 0.244 {\scriptsize[0.233, 0.254]} & 0.061 {\scriptsize[0.057, 0.061]} & 0.063 {\scriptsize[0.060, 0.069]} \\
     & $10^{5}$ & 0.280 {\scriptsize[0.268, 0.298]} & 0.060 {\scriptsize[0.059, 0.065]} & 0.070 {\scriptsize[0.065, 0.072]} \\
     & $10^{7}$ & 5.13 {\scriptsize[5.11, 5.14]} & 1.26 {\scriptsize[1.25, 1.26]} & 2.66 {\scriptsize[2.65, 2.69]} \\
    Poincar\'e $\log_0$ & $10^{3}$ & 0.174 {\scriptsize[0.170, 0.188]} & 0.056 {\scriptsize[0.054, 0.060]} & 0.071 {\scriptsize[0.070, 0.130]} \\
     & $10^{5}$ & 0.213 {\scriptsize[0.195, 0.222]} & 0.058 {\scriptsize[0.053, 0.059]} & 0.068 {\scriptsize[0.067, 0.073]} \\
     & $10^{7}$ & 3.88 {\scriptsize[3.88, 3.90]} & 1.25 {\scriptsize[1.24, 1.26]} & 1.95 {\scriptsize[1.91, 1.99]} \\
    Poincar\'e M\"obius $\oplus$ & $10^{3}$ & 0.063 {\scriptsize[0.061, 0.067]} & 0.061 {\scriptsize[0.060, 0.063]} & 0.071 {\scriptsize[0.062, 0.075]} \\
     & $10^{5}$ & 0.070 {\scriptsize[0.067, 0.159]} & 0.065 {\scriptsize[0.060, 0.144]} & 0.102 {\scriptsize[0.094, 0.121]} \\
     & $10^{7}$ & 1.82 {\scriptsize[1.82, 1.82]} & 1.82 {\scriptsize[1.82, 1.82]} & 4.64 {\scriptsize[4.60, 4.64]} \\
    Poincar\'e \texttt{ptransp} & $10^{3}$ & 0.061 {\scriptsize[0.059, 0.065]} & 0.064 {\scriptsize[0.063, 0.068]} & 0.066 {\scriptsize[0.064, 0.081]} \\
     & $10^{5}$ & 0.074 {\scriptsize[0.069, 0.114]} & 0.072 {\scriptsize[0.069, 0.119]} & 0.121 {\scriptsize[0.115, 0.151]} \\
     & $10^{7}$ & 2.39 {\scriptsize[2.37, 2.39]} & 2.39 {\scriptsize[2.37, 2.39]} & 5.83 {\scriptsize[5.82, 5.84]} \\
    Poincar\'e linear (fwd+bwd) & $10^{3}$ & n/a & n/a & 0.141 {\scriptsize[0.135, 0.171]} \\
     & $10^{5}$ & n/a & n/a & 0.293 {\scriptsize[0.281, 0.323]} \\
     & $10^{7}$ & n/a & n/a & 19.1 {\scriptsize[19.1, 19.2]} \\
    \midrule
    Hyperboloid \texttt{dist} & $10^{3}$ & 0.079 {\scriptsize[0.076, 0.082]} & n/a & 0.062 {\scriptsize[0.060, 0.064]} \\
     & $10^{5}$ & 0.096 {\scriptsize[0.093, 0.158]} & n/a & 0.113 {\scriptsize[0.110, 0.115]} \\
     & $10^{7}$ & 4.39 {\scriptsize[4.37, 10.3]} & n/a & 6.68 {\scriptsize[6.67, 6.70]} \\
    Hyperboloid $\exp_0$ & $10^{3}$ & 0.068 {\scriptsize[0.066, 0.071]} & n/a & 0.076 {\scriptsize[0.061, 0.123]} \\
     & $10^{5}$ & 0.127 {\scriptsize[0.125, 0.132]} & n/a & 0.110 {\scriptsize[0.108, 0.110]} \\
     & $10^{7}$ & 7.08 {\scriptsize[7.06, 7.41]} & n/a & 3.65 {\scriptsize[3.63, 3.69]} \\
    Hyperboloid $\log_0$ & $10^{3}$ & 0.077 {\scriptsize[0.074, 0.079]} & n/a & 0.067 {\scriptsize[0.060, 0.074]} \\
     & $10^{5}$ & 0.137 {\scriptsize[0.136, 0.146]} & n/a & 0.067 {\scriptsize[0.066, 0.093]} \\
     & $10^{7}$ & 8.20 {\scriptsize[8.18, 8.20]} & n/a & 2.78 {\scriptsize[2.73, 2.79]} \\
    Hyperboloid \texttt{ptransp} & $10^{3}$ & 0.112 {\scriptsize[0.106, 0.114]} & n/a & 0.090 {\scriptsize[0.064, 0.101]} \\
     & $10^{5}$ & 0.415 {\scriptsize[0.410, 0.436]} & n/a & 0.197 {\scriptsize[0.194, 0.234]} \\
     & $10^{7}$ & 33.2 {\scriptsize[33.2, 33.2]} & n/a & 16.0 {\scriptsize[16.0, 16.1]} \\
    Hyperboloid linear (fwd+bwd) & $10^{3}$ & n/a & n/a & 0.080 {\scriptsize[0.064, 0.102]} \\
     & $10^{5}$ & n/a & n/a & 0.140 {\scriptsize[0.138, 0.145]} \\
     & $10^{7}$ & n/a & n/a & 6.26 {\scriptsize[6.24, 6.27]} \\
      \bottomrule
    \end{tabular}
\end{apptable}

\begin{apptable}
    \centering
    \small
    \caption{Eager wall-clock time per call (ms) on the same H100 runs as Table~\ref{tab:speed}: geoopt and HypLL without \texttt{torch.compile}, hyperbolix as \texttt{jax.vmap} over the single-point API without \texttt{jax.jit}. The hyperbolix column measures host-side re-tracing on every call, which is nearly independent of batch size until the kernel dominates; it is not a kernel comparison. Aggregation and n/a semantics as in Table~\ref{tab:speed}.}
    \label{tab:speed_cuda_eager}
    \begin{tabular}{llccc}
      \toprule
      Operation & Batch & geoopt (eager) & HypLL (eager) & hyperbolix (eager) \\
      \midrule
    Poincar\'e \texttt{dist} & $10^{3}$ & 0.210 {\scriptsize[0.197, 0.275]} & 0.301 {\scriptsize[0.280, 0.418]} & 1.17 {\scriptsize[1.13, 1.63]} \\
     & $10^{5}$ & 0.245 {\scriptsize[0.231, 0.264]} & 0.313 {\scriptsize[0.304, 0.589]} & 1.60 {\scriptsize[1.56, 1.67]} \\
     & $10^{7}$ & 15.4 {\scriptsize[15.4, 15.4]} & 20.3 {\scriptsize[20.3, 20.4]} & 5.34 {\scriptsize[5.13, 5.88]} \\
    Poincar\'e $\exp_0$ & $10^{3}$ & 0.181 {\scriptsize[0.177, 0.322]} & 0.195 {\scriptsize[0.188, 0.335]} & 5.59 {\scriptsize[5.13, 6.91]} \\
     & $10^{5}$ & 0.216 {\scriptsize[0.211, 0.227]} & 0.212 {\scriptsize[0.206, 0.220]} & 5.84 {\scriptsize[5.28, 6.04]} \\
     & $10^{7}$ & 9.97 {\scriptsize[9.96, 9.98]} & 11.3 {\scriptsize[11.3, 11.3]} & 7.95 {\scriptsize[7.03, 9.98]} \\
    Poincar\'e $\log_0$ & $10^{3}$ & 0.134 {\scriptsize[0.106, 0.163]} & 0.078 {\scriptsize[0.061, 0.093]} & 2.69 {\scriptsize[2.50, 3.54]} \\
     & $10^{5}$ & 0.127 {\scriptsize[0.121, 0.132]} & 0.077 {\scriptsize[0.076, 0.077]} & 2.62 {\scriptsize[2.60, 2.74]} \\
     & $10^{7}$ & 3.86 {\scriptsize[3.84, 3.87]} & 5.00 {\scriptsize[5.00, 5.01]} & 4.49 {\scriptsize[3.77, 6.64]} \\
    Poincar\'e M\"obius $\oplus$ & $10^{3}$ & 0.203 {\scriptsize[0.179, 0.317]} & 0.228 {\scriptsize[0.208, 0.362]} & 19.8 {\scriptsize[19.5, 20.0]} \\
     & $10^{5}$ & 0.229 {\scriptsize[0.221, 0.237]} & 0.238 {\scriptsize[0.233, 0.533]} & 20.4 {\scriptsize[20.0, 20.7]} \\
     & $10^{7}$ & 19.0 {\scriptsize[19.0, 19.0]} & 16.8 {\scriptsize[16.8, 16.8]} & 33.6 {\scriptsize[33.5, 33.7]} \\
    Poincar\'e \texttt{ptransp} & $10^{3}$ & 0.208 {\scriptsize[0.198, 0.291]} & 0.487 {\scriptsize[0.468, 0.688]} & 23.8 {\scriptsize[23.7, 24.5]} \\
     & $10^{5}$ & 0.312 {\scriptsize[0.307, 0.627]} & 0.512 {\scriptsize[0.499, 0.999]} & 25.1 {\scriptsize[25.0, 25.7]} \\
     & $10^{7}$ & 24.7 {\scriptsize[24.7, 24.7]} & 38.6 {\scriptsize[38.6, 38.6]} & 45.3 {\scriptsize[45.3, 45.5]} \\
    Poincar\'e linear (fwd+bwd) & $10^{3}$ & n/a & 3.26 {\scriptsize[2.13, 3.85]} & 39.1 {\scriptsize[39.0, 40.4]} \\
     & $10^{5}$ & n/a & 3.19 {\scriptsize[1.55, 4.86]} & 40.2 {\scriptsize[38.9, 40.7]} \\
     & $10^{7}$ & n/a & 95.8 {\scriptsize[95.7, 95.8]} & 116 {\scriptsize[114, 120]} \\
    \midrule
    Hyperboloid \texttt{dist} & $10^{3}$ & 0.183 {\scriptsize[0.176, 0.190]} & n/a & 1.62 {\scriptsize[1.57, 1.83]} \\
     & $10^{5}$ & 0.201 {\scriptsize[0.194, 0.212]} & n/a & 1.53 {\scriptsize[1.53, 1.58]} \\
     & $10^{7}$ & 18.3 {\scriptsize[18.3, 18.3]} & n/a & 10.5 {\scriptsize[10.2, 10.6]} \\
    Hyperboloid $\exp_0$ & $10^{3}$ & 0.262 {\scriptsize[0.238, 0.271]} & n/a & 12.4 {\scriptsize[9.29, 15.2]} \\
     & $10^{5}$ & 0.188 {\scriptsize[0.183, 0.194]} & n/a & 13.9 {\scriptsize[10.3, 14.9]} \\
     & $10^{7}$ & 18.4 {\scriptsize[18.4, 18.5]} & n/a & 17.9 {\scriptsize[17.2, 18.6]} \\
    Hyperboloid $\log_0$ & $10^{3}$ & 0.214 {\scriptsize[0.195, 0.308]} & n/a & 4.39 {\scriptsize[3.74, 6.32]} \\
     & $10^{5}$ & 0.250 {\scriptsize[0.241, 0.278]} & n/a & 5.95 {\scriptsize[5.23, 6.29]} \\
     & $10^{7}$ & 19.6 {\scriptsize[19.5, 19.6]} & n/a & 8.28 {\scriptsize[8.22, 9.72]} \\
    Hyperboloid \texttt{ptransp} & $10^{3}$ & 0.421 {\scriptsize[0.411, 0.449]} & n/a & 42.1 {\scriptsize[39.6, 44.2]} \\
     & $10^{5}$ & 0.624 {\scriptsize[0.613, 0.784]} & n/a & 45.3 {\scriptsize[41.8, 46.4]} \\
     & $10^{7}$ & 61.4 {\scriptsize[61.4, 61.5]} & n/a & 79.2 {\scriptsize[79.1, 81.9]} \\
    Hyperboloid linear (fwd+bwd) & $10^{3}$ & n/a & n/a & 12.6 {\scriptsize[12.2, 12.7]} \\
     & $10^{5}$ & n/a & n/a & 12.6 {\scriptsize[12.5, 13.0]} \\
     & $10^{7}$ & n/a & n/a & 31.9 {\scriptsize[31.1, 32.8]} \\
      \bottomrule
    \end{tabular}
\end{apptable}

\subsection{Speed on a CPU}
\label{app:speed_cpu}

The CPU benchmark ran on a two-socket Intel Xeon Silver 4214R (12 cores and 24 hardware threads per socket, 48 logical CPUs) with Python 3.13.8, JAX 0.11.0, and PyTorch 2.13.0 executing on the CPU, over batch sizes $10^2$ to $10^5$; the tables show $10^5$. The machine was shared, though more lightly than in some other launches of this benchmark: the one-minute load average immediately before the first of the five default-thread launches was 0.84.

A CPU comparison between these libraries is a comparison between two runtimes' threading defaults. On this host PyTorch's ATen and Inductor kernels use 24 threads (\texttt{torch.get\_num\_threads()}), and in a three-launch probe at batch $10^5$ outside the released tables, the torch columns kept about 24 cores busy (CPU time over wall time) while XLA:CPU kept about 7--8; no \texttt{XLA\_FLAGS} setting resizes XLA's pool in jaxlib 0.11, and \texttt{OMP\_NUM\_THREADS} moves only PyTorch. We therefore report two tables. Table~\ref{tab:speed_cpu_default} is each runtime at its default; Table~\ref{tab:speed_cpu_onethread} and Figure~\ref{fig:speed_cpu_onethread} are a \emph{one-core control}: five further launches of the unmodified script under \texttt{taskset -c 2} with \texttt{OMP\_NUM\_THREADS=MKL\_NUM\_THREADS=1}, which pins all three libraries to one physical core (both the core and its hyperthread sibling were 99--100\% idle in the three seconds sampled just before the first launch, one-minute load average 3.38 at that moment) and holds PyTorch to one thread. \texttt{taskset} is the only knob that moves both runtimes, and each JSON records the affinity mask and thread count. The control is the like-for-like comparison; the default table is what a user sees without tuning.

At one core, batch $10^5$, compiled against compiled, hyperbolix jit is fastest on four of the nine primitive rows. On three of them the five-launch brackets are disjoint: Poincar\'e $\log_0$ (1.9$\times$ over geoopt, 2.2$\times$ over HypLL), hyperboloid $\exp_0$ (1.5$\times$ over geoopt), and hyperboloid $\log_0$ (2.2$\times$). On the fourth, Poincar\'e $\exp_0$, it is 1.6$\times$ faster than geoopt but only 1.06$\times$ faster than HypLL, with overlapping brackets, so that cell is a tie with HypLL. It loses the other five: HypLL leads on Poincar\'e distance (2.6$\times$) and M\"obius addition (4.6$\times$; geoopt 3.8$\times$); geoopt and HypLL are within noise of each other on Poincar\'e parallel transport (8.64 versus 8.58\,ms) and both are 3.0$\times$ faster than hyperbolix there; and geoopt's compiled kernels lead on hyperboloid distance (1.8$\times$) and hyperboloid parallel transport (1.4$\times$), where HypLL has no cells. The default-table gaps are larger (up to ${\sim}12\times$ on Poincar\'e M\"obius addition) because PyTorch uses 24 threads against XLA's ${\sim}8$. The one-core control shrinks them to the ratios above, so most, though not all, of the default-table gap is thread-count asymmetry.

The default table carries its own caveats. Its per-column across-launch max/min at $10^5$ is 3.89, 3.74, 3.33, 1.26, 1.52, and 1.59 for geoopt eager, geoopt compiled, HypLL eager, HypLL compiled, hyperbolix eager, and hyperbolix jit; the widest hyperbolix jit cell is Poincar\'e M\"obius addition, one launch at 13.2\,ms against four at 8.3--9.3\,ms, rather than systematic. We did not run a same-day control launch for this table, so we do not quote any single cell to a precision tighter than the reported five-launch bracket. Finally, the CPU and CUDA runs differ in machine, kernel, Python, JAX, and PyTorch build, and the CPU sweep stops at $10^5$, so CPU against CUDA is not a single-variable comparison.

\begin{appfigure}
  \centering
  \includegraphics[width=\columnwidth]{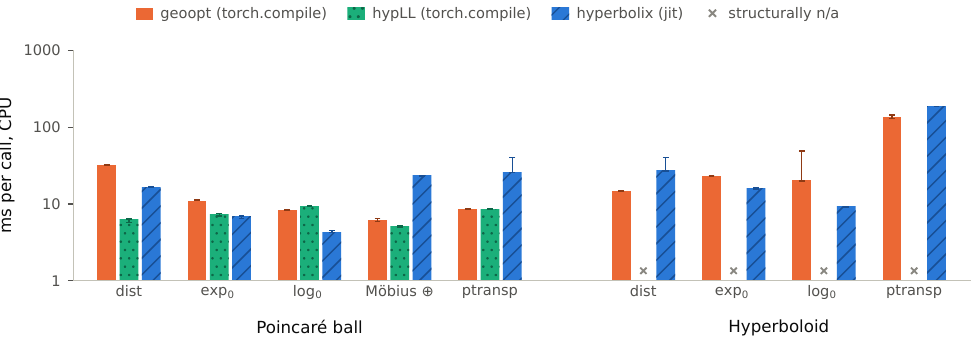}
  \caption{Compiled wall-clock time per call on one CPU core per library (Table~\ref{tab:speed_cpu_onethread}), batch $10^5$, $D = 32$, float32, primitives only; the layer rows and the eager columns are table-only. Bars are the median of five launch medians, error bars the min--max across launches. The axis is logarithmic with a floor at 1\,ms, so bar tops, not bar lengths, are proportional to time. The four hyperboloid HypLL slots are structurally absent and drawn as grey crosses.}
  \label{fig:speed_cpu_onethread}
\end{appfigure}

\begin{apptable}
    \centering
    \footnotesize
    \setlength{\tabcolsep}{3.5pt}
    \caption{One-core control: wall-clock time per call (ms) at batch $10^5$, float32, $D = 32$, with all three libraries pinned to one physical core of the Xeon Silver 4214R (\texttt{taskset -c 2}) and PyTorch held to one thread (\texttt{OMP\_NUM\_THREADS=MKL\_NUM\_THREADS=1}). Cell = median of five launch medians; bracket = min--max across launches. The widest brackets are geoopt's hyperboloid $\log_0$ (max/min 2.4$\times$ compiled, 1.6$\times$ eager: two launches at ${\sim}48$\,ms against three at ${\sim}20$\,ms) and three cells where a single launch departs from the other four: hyperbolix jit on Poincar\'e parallel transport (1.6$\times$) and hyperboloid distance (1.5$\times$), and geoopt eager on Poincar\'e parallel transport (1.5$\times$). Hyperboloid parallel transport is among the tightest cells (hyperbolix jit 1.006$\times$). Cause unmeasured. n/a semantics as in Table~\ref{tab:speed}.}
    \label{tab:speed_cpu_onethread}
    \resizebox{\textwidth}{!}{%
    \begin{tabular}{lcccccc}
      \toprule
      Operation & geoopt (eager) & geoopt (compile) & HypLL (eager) & HypLL (compile) & hyperbolix (eager) & hyperbolix (jit) \\
      \midrule
    Poincar\'e \texttt{dist} & 32.2 {\scriptsize[32.0, 32.6]} & 32.6 {\scriptsize[32.3, 32.9]} & 29.6 {\scriptsize[29.6, 29.9]} & 6.26 {\scriptsize[5.77, 6.35]} & 39.6 {\scriptsize[39.4, 39.7]} & 16.4 {\scriptsize[16.4, 16.6]} \\
    Poincar\'e $\exp_0$ & 20.5 {\scriptsize[20.5, 20.8]} & 11.1 {\scriptsize[11.0, 11.2]} & 49.7 {\scriptsize[49.5, 50.0]} & 7.23 {\scriptsize[6.95, 7.58]} & 12.3 {\scriptsize[12.1, 12.5]} & 6.82 {\scriptsize[6.49, 6.97]} \\
    Poincar\'e $\log_0$ & 8.06 {\scriptsize[8.00, 8.13]} & 8.33 {\scriptsize[8.27, 8.39]} & 7.54 {\scriptsize[7.36, 7.54]} & 9.35 {\scriptsize[9.31, 9.42]} & 9.96 {\scriptsize[9.87, 10.00]} & 4.29 {\scriptsize[4.25, 4.49]} \\
    Poincar\'e M\"obius $\oplus$ & 63.5 {\scriptsize[63.3, 64.0]} & 6.13 {\scriptsize[5.92, 6.43]} & 50.9 {\scriptsize[50.8, 51.2]} & 5.06 {\scriptsize[5.03, 5.15]} & 112 {\scriptsize[93.0, 130]} & 23.4 {\scriptsize[23.2, 23.6]} \\
    Poincar\'e \texttt{ptransp} & 80.9 {\scriptsize[54.1, 81.2]} & 8.64 {\scriptsize[8.57, 8.77]} & 86.8 {\scriptsize[63.4, 86.8]} & 8.58 {\scriptsize[8.56, 8.76]} & 136 {\scriptsize[130, 137]} & 26.1 {\scriptsize[25.7, 40.3]} \\
    Poincar\'e linear (fwd+bwd) & n/a & n/a & 413 {\scriptsize[316, 421]} & n/a & 336 {\scriptsize[313, 348]} & 181 {\scriptsize[181, 181]} \\
    \midrule
    Hyperboloid \texttt{dist} & 25.2 {\scriptsize[25.0, 25.7]} & 14.7 {\scriptsize[14.7, 14.9]} & n/a & n/a & 29.0 {\scriptsize[28.8, 42.8]} & 26.9 {\scriptsize[26.9, 40.3]} \\
    Hyperboloid $\exp_0$ & 30.6 {\scriptsize[26.3, 30.8]} & 23.4 {\scriptsize[23.3, 23.6]} & n/a & n/a & 42.2 {\scriptsize[41.3, 47.8]} & 15.9 {\scriptsize[15.8, 16.3]} \\
    Hyperboloid $\log_0$ & 34.9 {\scriptsize[33.8, 53.3]} & 20.5 {\scriptsize[20.0, 48.7]} & n/a & n/a & 17.3 {\scriptsize[17.2, 17.8]} & 9.20 {\scriptsize[9.10, 9.23]} \\
    Hyperboloid \texttt{ptransp} & 149 {\scriptsize[145, 159]} & 136 {\scriptsize[130, 144]} & n/a & n/a & 261 {\scriptsize[234, 266]} & 186 {\scriptsize[186, 187]} \\
    Hyperboloid linear (fwd+bwd) & n/a & n/a & n/a & n/a & 77.0 {\scriptsize[74.9, 78.1]} & 25.2 {\scriptsize[24.9, 25.3]} \\
      \bottomrule
    \end{tabular}}
\end{apptable}

\begin{apptable}
    \centering
    \footnotesize
    \setlength{\tabcolsep}{3.5pt}
    \caption{Runtime-default threading: wall-clock time per call (ms) at batch $10^5$, float32, $D = 32$, on the same Xeon Silver 4214R with PyTorch at its default 24 threads and XLA:CPU at its default pool (one-minute load 0.84 at launch). Cell = median of five launch medians; bracket = min--max across launches; the brackets mix co-tenancy with launch-to-launch state. n/a semantics as in Table~\ref{tab:speed}.}
    \label{tab:speed_cpu_default}
    \resizebox{\textwidth}{!}{%
    \begin{tabular}{lcccccc}
      \toprule
      Operation & geoopt (eager) & geoopt (compile) & HypLL (eager) & HypLL (compile) & hyperbolix (eager) & hyperbolix (jit) \\
      \midrule
    Poincar\'e \texttt{dist} & 5.62 {\scriptsize[5.30, 6.00]} & 5.72 {\scriptsize[5.69, 6.19]} & 5.69 {\scriptsize[5.14, 5.80]} & 0.949 {\scriptsize[0.879, 1.10]} & 24.5 {\scriptsize[24.0, 25.6]} & 10.1 {\scriptsize[9.87, 10.8]} \\
    Poincar\'e $\exp_0$ & 3.56 {\scriptsize[3.35, 13.0]} & 2.22 {\scriptsize[1.85, 2.77]} & 13.1 {\scriptsize[12.4, 13.4]} & 1.22 {\scriptsize[1.21, 1.40]} & 6.05 {\scriptsize[5.96, 6.45]} & 2.49 {\scriptsize[2.44, 2.63]} \\
    Poincar\'e $\log_0$ & 1.65 {\scriptsize[1.55, 1.74]} & 1.80 {\scriptsize[1.71, 1.93]} & 1.10 {\scriptsize[1.00, 1.23]} & 0.735 {\scriptsize[0.669, 0.780]} & 4.02 {\scriptsize[3.89, 4.34]} & 1.91 {\scriptsize[1.87, 1.92]} \\
    Poincar\'e M\"obius $\oplus$ & 14.2 {\scriptsize[13.9, 14.3]} & 0.844 {\scriptsize[0.758, 0.853]} & 4.30 {\scriptsize[3.85, 12.8]} & 0.719 {\scriptsize[0.620, 0.772]} & 46.1 {\scriptsize[37.8, 57.4]} & 8.86 {\scriptsize[8.33, 13.2]} \\
    Poincar\'e \texttt{ptransp} & 6.99 {\scriptsize[6.09, 7.80]} & 1.08 {\scriptsize[1.00, 1.10]} & 8.15 {\scriptsize[7.98, 8.63]} & 1.10 {\scriptsize[1.03, 1.13]} & 53.5 {\scriptsize[50.7, 53.8]} & 7.09 {\scriptsize[7.01, 7.25]} \\
    Poincar\'e linear (fwd+bwd) & n/a & n/a & 86.1 {\scriptsize[75.9, 89.2]} & n/a & 143 {\scriptsize[128, 150]} & 80.8 {\scriptsize[80.1, 81.2]} \\
    \midrule
    Hyperboloid \texttt{dist} & 4.59 {\scriptsize[4.34, 4.85]} & 1.93 {\scriptsize[1.49, 2.01]} & n/a & n/a & 8.85 {\scriptsize[8.69, 9.10]} & 6.12 {\scriptsize[6.04, 6.14]} \\
    Hyperboloid $\exp_0$ & 5.66 {\scriptsize[4.98, 6.68]} & 7.63 {\scriptsize[7.50, 8.29]} & n/a & n/a & 19.7 {\scriptsize[18.5, 20.3]} & 9.88 {\scriptsize[9.53, 10.1]} \\
    Hyperboloid $\log_0$ & 18.5 {\scriptsize[6.90, 19.4]} & 30.4 {\scriptsize[8.66, 32.4]} & n/a & n/a & 7.81 {\scriptsize[7.51, 8.39]} & 3.27 {\scriptsize[3.21, 3.29]} \\
    Hyperboloid \texttt{ptransp} & 41.7 {\scriptsize[39.6, 46.0]} & 88.6 {\scriptsize[86.0, 91.9]} & n/a & n/a & 111 {\scriptsize[100, 132]} & 61.4 {\scriptsize[58.5, 61.9]} \\
    Hyperboloid linear (fwd+bwd) & n/a & n/a & n/a & n/a & 31.4 {\scriptsize[29.8, 39.9]} & 6.78 {\scriptsize[6.64, 6.84]} \\
      \bottomrule
    \end{tabular}}
\end{apptable}

\subsection{Two-point cancellation on the hyperboloid}
\label{app:cancellation}

Every accuracy number so far comes from an operation anchored at the origin: $\exp_0$, $\log_0$, $\mathrm{dist}_0$, and $\mathrm{ptransp}$ from the origin. The hyperboloid's two-point Minkowski inner product $\langle x, y \rangle_{\mathcal{L}} = -x_0 y_0 + \langle x_s, y_s \rangle$ is different: away from the origin, both terms grow like $e^{a}$ for a point at geodesic radius $a$ (with $a = \sqrt{c}\,d(o, x)$), so their difference is a subtraction of two $O(e^{2a})$ quantities, a cancellation no origin-anchored operation exercises. We test it directly, with a construction that needs no oracle: in geodesic polar coordinates $\mathrm{polar}(r, u) = (\cosh(\sqrt{c}\,r)/\sqrt{c},\ \sinh(\sqrt{c}\,r)/\sqrt{c}\cdot u)$ for a unit direction $u$, two points on the same ray at radii $a$ and $a+1$ are exactly geodesic distance $1$ apart by construction, and the tangent vector $v = \bigl((\sinh a,\ \cosh a\,u) + (0, w)\bigr)/\sqrt{2}$ at the point at radius $a$, with $w$ a unit vector orthogonal to $u$. It has Riemannian norm exactly $1$ because it is the normalized sum of the ray's unit-speed velocity and a unit tangent orthogonal to it, and it points at $45^\circ$ to the ray. Ground truth is therefore a constant, not a transcribed formula, so no oracle-transcription error can enter; only the library's own arithmetic can produce a deviation. We sweep $a$ from $2$ to $16$ at $c = 1$, dimension $8$, $500$ sampled directions per cell, both dtypes, comparing hyperbolix against geoopt's \texttt{Lorentz} model. We evaluate the tangent norms for both libraries on the same set of vectors.

Table~\ref{tab:cancellation} reports the four radii cited by the accuracy claims below; the corresponding main-text figure (\texttt{results\_paper/hyperboloid\_cancellation.pdf}) plots the full sweep.  On the distance, both libraries' error grows like $e^{2a}$, consistent with the cancellation model, which makes this a slope claim rather than a pass/fail at one radius. They differ in the prefactor: geoopt computes the difference literally, and its float32 distance median passes 1\% error at $a \approx 7$ and returns \texttt{NaN} from $a = 9$ onward (its \texttt{arcosh} receives a negative argument once the cancellation consumes every significant digit). hyperbolix's cancellation-free form sits at the float32 resolution floor out to $a \approx 10$: through $a = 12$ its float32 maximum absolute deviation is $3.7 \times 10^{-4}$ (median $5.6 \times 10^{-5}$), and its float64 deviation is $\le 6.7 \times 10^{-16}$ (3~ulp) at every radius we tested, including $a = 16$. The model prediction is exact enough to check quantitatively: geoopt's float64 curve is its float32 curve shifted by $\Delta a = 10$, matching $\varepsilon_{64}/\varepsilon_{32} = 2^{-29}$ (median $3.43 \times 10^{-3}$ at $a = 16$ in float64, the same $3.43 \times 10^{-3}$ at $a = 6$ in float32).

hyperbolix is not unconditionally safe in float32; it leaves the resolution floor about five units of radius later than geoopt. Its own float32 median distance error reaches $7.2 \times 10^{-4}$ at $a = 14$ and $1.0 \times 10^{-1}$ at $a = 16$, both outside the table's printed range but visible in the main-text figure, which is why the sweep runs to $a = 16$ rather than stopping once geoopt has failed. Two hedges apply when quoting this comparison: above $a = 8$, geoopt's maximum (as opposed to its median) is reduction-order noise rather than a stable quantity, since every digit in a fully-cancelled float32 cell is meaningless -- quote its median, or simply "$O(1)$/\texttt{NaN}"; and the timed callables (Table~\ref{tab:cancellation} omits timing) show cost flat with radius for hyperbolix jit ($0.95$--$1.72$\,ms) and for geoopt eager ($0.25$--$0.81$\,ms) across $a = 2$ to $16$, a single launch with unequal thread pools between the two libraries, so it answers only "does cost grow with radius" and is not a speed claim in either direction.

  \begin{apptable}
      \centering
      \small
      \caption{Two-point Minkowski-inner-product cancellation on the hyperboloid, $c = 1$, dimension 8, 500 sampled directions per cell: relative max$|$dev$|$ (median$|$dev$|$) against the exact constructions of Section~\ref{app:cancellation}. Both true
  values are $1$, so relative and absolute deviation coincide. hyperbolix 1.3.0 against geoopt 0.5.1's \texttt{Lorentz} model. geoopt's distance is fully cancelled from $a = 9$ (\texttt{NaN}); its maxima at $a \ge 8$ are reduction-order noise and only
  its medians are meaningful. hyperbolix's distance stays at float32 rounding level through $a = 10$ and within $6.7\times10^{-16}$ in float64 at every radius; its tangent norm stays within $1.35\times$ of the input-rounding error from $a = 6$. Script:
  \texttt{scripts/hyperboloid\_cancellation.py}, CPU backend.}
      \label{tab:cancellation}
      \resizebox{\textwidth}{!}{%
      \begin{tabular}{llcccc}
        \toprule
        & & \multicolumn{2}{c}{dist} & \multicolumn{2}{c}{tangent norm} \\
        \cmidrule(lr){3-4} \cmidrule(lr){5-6}
        $a$ & dtype & hyperbolix & geoopt & hyperbolix & geoopt \\
        \midrule
      6 & float32 & $2.38\times10^{-7}$ ($5.96\times10^{-8}$) & $1.69\times10^{-2}$ ($3.43\times10^{-3}$) & $6.56\times10^{-6}$ ($7.15\times10^{-7}$) & $2.93\times10^{-3}$ ($9.76\times10^{-4}$) \\
      6 & float64 & $6.66\times10^{-16}$ ($0$) & $6.24\times10^{-11}$ ($1.29\times10^{-11}$) & $1.67\times10^{-14}$ ($1.78\times10^{-15}$) & $7.28\times10^{-12}$ ($1.82\times10^{-12}$) \\
      \midrule
      8 & float32 & $3.58\times10^{-7}$ ($0$) & $1.00\times10^{0}$ ($3.76\times10^{-2}$) & $4.73\times10^{-5}$ ($4.98\times10^{-6}$) & $1.34\times10^{-1}$ ($6.46\times10^{-2}$) \\
      8 & float64 & $4.44\times10^{-16}$ ($2.22\times10^{-16}$) & $2.70\times10^{-9}$ ($4.71\times10^{-10}$) & $1.13\times10^{-13}$ ($1.20\times10^{-14}$) & $3.49\times10^{-10}$ ($1.16\times10^{-10}$) \\
      \midrule
      10 & float32 & $2.50\times10^{-6}$ ($2.38\times10^{-7}$) & NaN (NaN) & $4.07\times10^{-4}$ ($3.81\times10^{-5}$) & $1.83\times10^{0}$ ($1.00\times10^{0}$) \\
      10 & float64 & $2.22\times10^{-16}$ ($0$) & $1.58\times10^{-7}$ ($4.48\times10^{-8}$) & $8.80\times10^{-13}$ ($8.15\times10^{-14}$) & $1.86\times10^{-8}$ ($3.73\times10^{-9}$) \\
      \midrule
      12 & float32 & $2.89\times10^{-4}$ ($4.53\times10^{-5}$) & NaN (NaN) & $2.52\times10^{-3}$ ($2.64\times10^{-4}$) & $2.67\times10^{1}$ ($1.50\times10^{1}$) \\
      12 & float64 & $4.44\times10^{-16}$ ($2.22\times10^{-16}$) & $7.56\times10^{-6}$ ($1.07\times10^{-6}$) & $7.37\times10^{-12}$ ($7.15\times10^{-13}$) & $7.15\times10^{-7}$ ($2.38\times10^{-7}$) \\
        \bottomrule
      \end{tabular}}
  \end{apptable}

\end{document}